\documentclass{article}

\usepackage[preprint]{neurips_2025}

\usepackage[utf8]{inputenc} %
\usepackage[T1]{fontenc}    %
\usepackage{hyperref}       %
\usepackage{url}            %
\usepackage{booktabs}       %
\usepackage{amsfonts}       %
\usepackage{enumitem}       %
\usepackage{nicefrac}       %
\usepackage{microtype}      %
\usepackage{xcolor}         %
\usepackage[normalem]{ulem}

\usepackage[legacy]{csvsimple}
\usepackage{multirow}
\usepackage{makecell}
\usepackage{arydshln}
\usepackage{tabularx}
\usepackage{algorithm}
\usepackage[linesnumbered,ruled,vlined,algo2e]{algorithm2e}
\usepackage{algpseudocode}
\usepackage{listings} 
\usepackage{amsfonts}
\usepackage{wrapfig}

\usepackage{graphicx}
\usepackage{amsmath}

\usepackage{duckuments}

\usepackage{subcaption}
\usepackage{caption}

\usepackage{float}

\newcommand{\review}[1]{{\color{black}#1}}

\newif\ifdraft
\drafttrue
\ifdraft
\newcommand{\gpc}[1]{{\color{blue}[\textbf{Gaurav:} #1]}}
\newcommand{\kac}[1]{{\color{purple}[\textbf{Kfir:} #1]}}
\newcommand{\opc}[1]{{\color{red}[\textbf{Or:} #1]}}
\newcommand{\dcc}[1]{{\color{orange}[\textbf{Danny:} #1]}}
\newcommand{\jacksoncomment}[1]{{\color{teal}[\textbf{Jackson:} #1]}}
\newcommand{\dosc}[1]{{\color{magenta}[\textbf{Daniil:} #1]}}

\newcommand{\jackson}[1]{{\color{teal}#1}}
\newcommand{\dos}[1]{{\color{magenta}#1}}

\else
\newcommand{\gpc}[1]{}
\newcommand{\kac}[1]{}
\newcommand{\opc}[1]{}
\newcommand{\dcc}[1]{}
\newcommand{\jacksoncomment}[1]{}
\newcommand{\dosc}[1]{}

\newcommand{\jackson}[1]{}
\newcommand{\dos}[1]{}

\fi

\newcommand{\refsec}[1]{Section~\ref{sec:#1}}

\newcommand{\refeq}[1]{Equation~\ref{eq:#1}}

\newcommand{\myparagraph}[1]{\vspace{-8pt}\paragraph{#1}}

\def\x{{\mathbf{x}}}
\def\y{{\mathbf{y}}}
\def\c{{\mathbf{c}}}

\def\z{{\mathbf{z}}}

\def\u{{\mathbf{u}}}
\def\b{{\mathbf{b}}}

\definecolor{plt_blue}{HTML}{1f77b4}
\definecolor{plt_orange}{HTML}{ff7f0e}
\definecolor{plt_green}{HTML}{2ca02c}
\definecolor{plt_purple}{HTML}{9467bd}
\definecolor{plt_red}{HTML}{d62728}

\title{Scaling Group Inference for \\ Diverse and High-Quality Generation}

\author{
  Gaurav Parmar$^{1}$ \quad
  Or Patashnik$^{2, 3}$ \quad
  Daniil Ostashev$^{2}$ \quad
  Kuan-Chieh Wang$^{2}$ \quad
  Kfir Aberman$^{2}$ \\
  \And
  Srinivasa Narasimhan$^{1}$ \quad
  Jun-Yan Zhu$^{1}$ \\ \\
  $^{1}$Carnegie Mellon University \quad $^{2}$Snap Research \quad $^{3}$Tel Aviv University
}

\begin{document}

\maketitle

\begin{abstract}
    Generative models typically sample outputs independently, and recent inference-time guidance and scaling algorithms focus on improving the quality of individual samples.
    However, in real-world applications, users are often presented with \emph{a set} of multiple images (e.g., 4-8) for each prompt, where independent sampling tends to lead to redundant results, limiting user choices and hindering idea exploration. 
     In this work, we introduce a scalable group inference method that improves both the diversity and quality of a group of samples. 
    We formulate group inference as a quadratic integer assignment problem: candidate outputs are modeled as graph nodes, and a subset is selected to optimize sample quality (unary term) while maximizing group diversity (binary term). 
    To substantially improve runtime efficiency, we progressively prune the candidate set using intermediate predictions, allowing our method to scale up to large candidate sets. 
    Extensive experiments show that our method significantly improves group diversity and quality compared to independent sampling baselines and recent inference algorithms.
    Our framework generalizes across a wide range of tasks, including text-to-image, image-to-image, image prompting, and video generation, enabling generative models to treat multiple outputs as cohesive groups rather than independent samples.  
\end{abstract}

\begin{figure*}[t]
    \centering 
    \includegraphics[width=\linewidth]{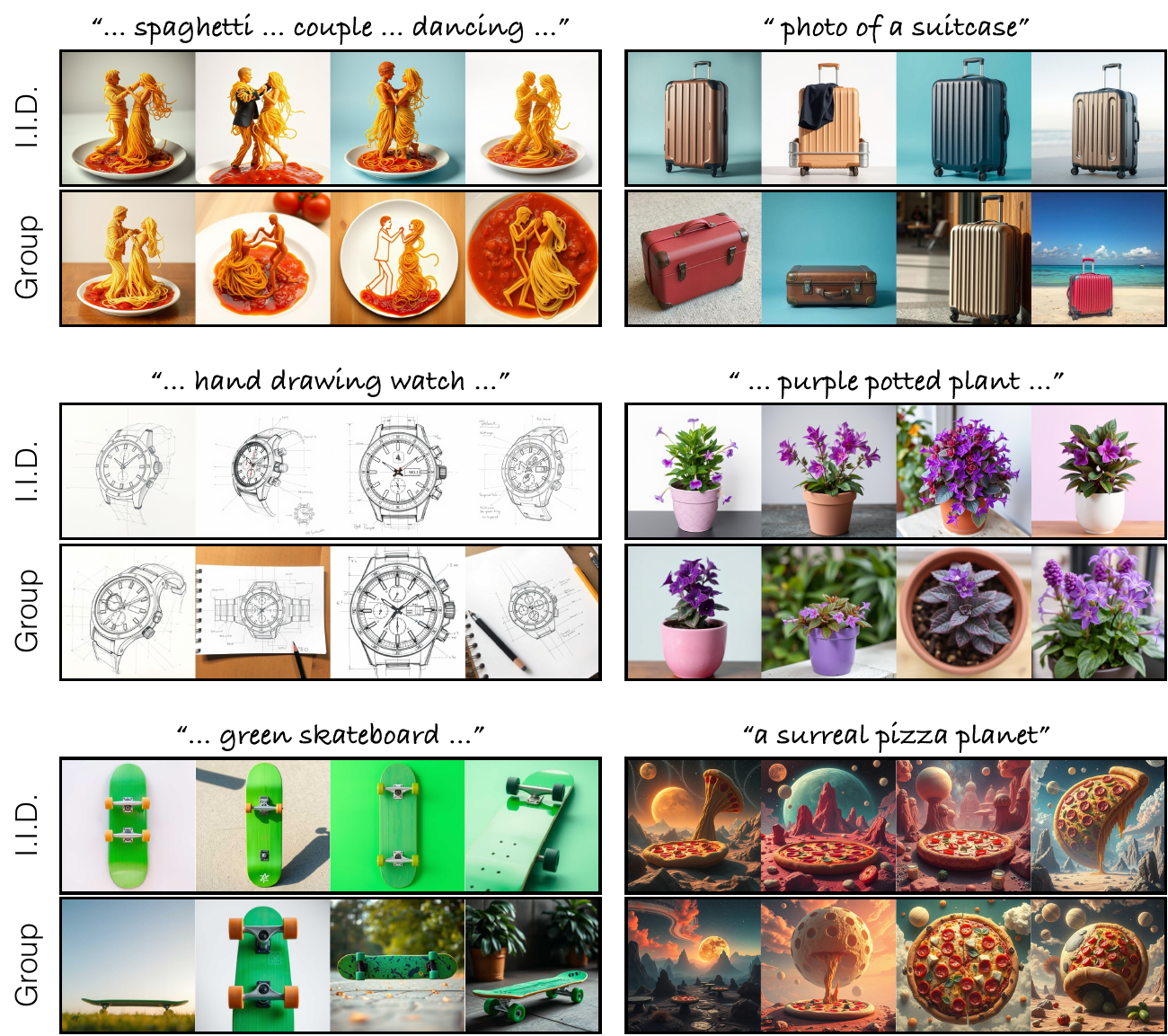}
    \caption{\textbf{Scalable Group Inference Results.} We show the advantage of our proposed group inference method over I.I.D. sampling. While I.I.D. sampling often yields repetitive results for the same prompt, our method generates a more diverse and high-quality collection of outputs. 
Please see our project \href{https://www.cs.cmu.edu/~group-inference/}{website} for more results. 
    } 
    \label{fig:teaser}
    \vspace{-3mm}
\end{figure*}

\section{Introduction} \label{sec:intro}
Recent advances in generative models, such as diffusion models, have driven significant efforts in inference-time guidance and scaling techniques~\cite{ho2021classifier,ma2025inference,parmar2023zero}. These methods effectively improve various aspects of output quality, such as alignment with text prompts or image aesthetics, and offer fine-grained controls over the output.  However, much recent work primarily focuses on enhancing the quality of \emph{individual} samples generated in isolation.

Yet, in real-world applications, users are often shown a group of samples rather than just one. For example, many text-to-image platforms~\cite{MidjourneyWebsite,AdobeFirefly}  display a grid of four to eight images per prompt by default, a practice that offers users crucial benefits: more diverse choices regarding layout, lighting, and style, and new inspirations and ideas for prompt refinement and local edits. This creates a gap between current research, focused on independent samples, and the practical need for diverse, high-quality groups in content creation workflows. %
How can we close this gap?

In this work, we propose a \emph{scalable group inference} method to jointly improve the diversity and quality of a collection of generated samples. We formulate this task as a quadratic integer programming problem, representing output candidates as graph nodes. From a large set of $M$ candidates, we select a subset of size $K$ that maximizes a combination of individual sample quality, as a unary term,  and group diversity as a binary term. However, a direct approach involves running the $T$-step denoising process for all $M$ candidates, resulting in an $\mathcal{O}(MT)$ complexity. This is computationally expensive for large $M$ and $T$ (e.g., M=128, T=20). To address this, we introduce an efficient progressive selection strategy that leverages intermediate predictions during denoising to iteratively prune the candidate set. 
This approach is grounded in the insight that these intermediate predictions of the final output, despite originating from a long denoising chain, serve as effective \emph{previews} of the final image at each step (Figure~\ref{fig:correlations}).
This approach achieves a complexity of $\mathcal{O}(M + K T)$, where $K=4$, enabling us to scale up our group inference to handle large candidate sets. 

Extensive experiments have shown that our group inference method significantly outperforms independent sampling baselines and recent single-sample inference algorithms across various generative tasks and modalities, including text-to-image, image-to-image, image prompting, and video generation.  Our method scales much better and produces more diverse and realistic outputs given the same compute budget. We further provide a comprehensive ablation study demonstrating the effectiveness of our design choices. Our framework enables generative models to treat multiple outputs as cohesive groups, aligning more closely with real-world workflows. 
In summary, our contributions are:. 

\begin{itemize}[leftmargin=12pt]
\setlength{\parskip}{0pt}
\item We propose a new, scalable group inference algorithm by selecting a group of K samples from M candidates as a quadratic integer programming problem to maximize sample quality and group diversity.
\item We introduce a progressive pruning strategy to further scale our method. Our technique uses intermediate $\x_0$ predictions as previews to iteratively prune candidates, reducing complexity from $\mathcal{O}(MT)$ to  $\mathcal{O}(M + K T)$, where $K $ is much smaller than $M$. 
\item Extensive evaluation on text-to-image, image-to-image, image prompting, and video generation shows our method outperforms baselines, producing more diverse and realistic outputs within similar cost budgets.
\end{itemize}

\section{Related Works} \label{sec:related}
\myparagraph{Diffusion models.} are a powerful class of generative models that synthesize high-quality samples through iterative denoising~\cite{sohl2015deep, ho2020denoising,song2020score}. Successful in text-to-image synthesis \cite{rombach2022high, podell2024sdxl, saharia2022photorealistic}, their application later extends to video \cite{ho2022video, blattmann2023align, blattmann2023stable} and 3D synthesis \cite{poole2022dreamfusion, lin2023magic3d, shi2023mvdream}. However, common strategies to improve individual quality, such as fine-tuning for high quality,  less diverse datasets~\cite{rombach2022high} or strong classifier-free guidance (CFG) \cite{ho2021classifier}, often sacrifice diversity \cite{astolfi2024consistency}. Additional conditioning methods like spatial controls \cite{zhang2023adding} or image prompting \cite{ye2023ipadapter} improve controllability but also reduce diversity, especially with strong guidance values. This lack of diversity is further worsened by one-step or few-step generators \cite{sauer2024fast, sauer2024adversarial, kang2024distilling, yin2024one}. Our work addresses this trade-off with group inference, enhancing both sample quality and diversity in batches, and demonstrating applicability across various controls (text, spatial, visual prompts).

\myparagraph{Diffusion Inference and Guidance.} 
Inference-time guidance effectively improves sample quality and controllability of diffusion models without costly model finetuning. Early methods, such as classifier guidance \cite{dhariwal2021diffusion} and widely-used classifier-free guidance (CFG) \cite{ho2021classifier}, significantly increase sample quality, often at the cost of diversity. Recent approaches manipulate internal representations, such as cross-attention maps \cite{parmar2023zero}, or incorporate spatial control from inputs such as layouts or sketches \cite{chen2024training, kim2023dense, phung2024grounded, voynov2023sketch, he2023localized}. Other strategies apply guidance over limited intervals \cite{kynkaanniemi2024applying} or thresholding to CFG to reduce saturation~\cite{saharia2022photorealistic}.

While the above techniques focus on improving individual samples, our group inference approach explicitly optimizes collective properties, balancing single-sample quality and inter-sample diversity. A closely related work is particle guidance~\cite{corso2023particle}, which incorporates a pairwise potential during denoising steps to encourage diversity. Our method differs in three ways. First, our method improves both quality and diversity, while particle guidance often hurts image quality, as shown in experiments (\refsec{baselines_tradeoff}). Second, our method scales effectively to a large number of images through early candidate pruning and sample selection, avoiding expensive optimization. In contrast, particle guidance is limited to small sets (e.g., four images) due to memory-intensive gradient computation of the pairwise terms. Third, our framework supports non-differentiable quality and diversity terms, enabling the use of metrics derived from multimodal LLMs.

\myparagraph{Inference-time Scaling.} Test-time scaling, leveraging methods like chain-of-thought~\cite{wei2022chain}, proposer and verifier~\cite{snell2024scaling}, or multi-step reasoning, has become a key research area for large-language models~\cite{muennighoff2025s1}. The idea is to increase inference-time computation in exchange for improved performance from a pre-trained model. 
Recently, researchers have adopted the inference-time scaling for diffusion models~\cite{ma2025inference}, which uses off-the-shelf models and evaluation metrics to search for better noises and increase the sample quality, often requiring thousands to tens of thousands of function evaluations (NFEs). 
However, text-to-image models differ from LLMs in three ways:  they are often more computationally expensive~\cite{li2024svdqunat}, users often pay 5 to 10 cents per image on leading platforms, and users demand low latency.  In our work, we show that our test-time scaling method balances the computational cost and quality and diversity improvement. %

\begin{figure*}[t!]
    \centering 
    \includegraphics[width=\linewidth]{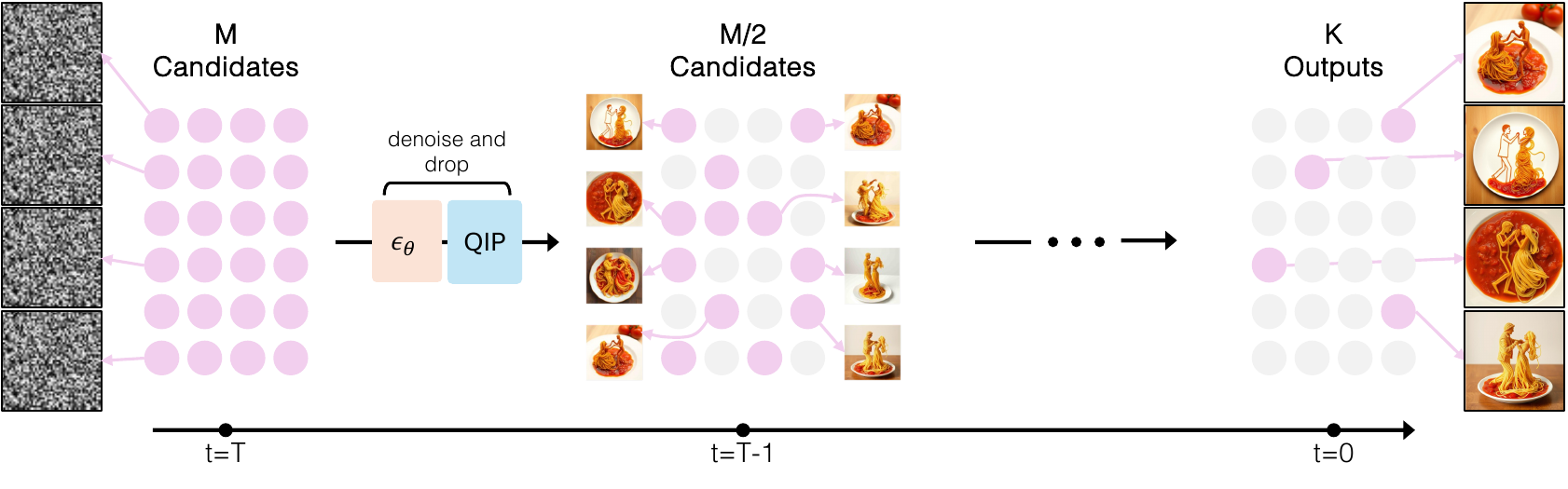}
    \caption{ \textbf{Overview.} Given a large number of $M$ candidate noises, we gradually reduce the number of candidate sets through iterative denoising and pruning steps. At each step, we first leverage the diffusion model $\epsilon_{\theta}$ to denoise the sample. We then compute the quality metric (unary term) and pairwise distances (binary term), and solve a quadratic integer programming (QIP) program to progressively prune the candidate set,  yielding a final group of $K$ diverse and high-quality outputs.
    } 
    \vspace{-10pt}
    \label{fig:method}
\end{figure*}

\section{Method} \label{sec:method}

We propose \emph{Scalable Group Inference}, a test-time selection framework that chooses a diverse, high-quality subset from a large pool of generated outputs. The method relies on a scoring objective that combines a unary term that measures the quality of an individual sample and a binary term that computes pairwise properties such as image distances. We first formulate this as a quadratic integer programming problem (QIP) over binary selection variables. Then, to reduce compute cost, we introduce a progressive filtering strategy that prunes low-quality candidates early using intermediate predictions from partially denoised samples. We now describe both components in detail.

\subsection{Formulation}
\label{sec:formulation}

Given a generative model $G_\theta(\z, \c)$ that maps latent noise $\z \sim p(\z)$ and condition $\c$ to outputs $\x$, our goal is to obtain a \emph{set} of $K$ outputs, $\{\x^{(i)}\}_{i=1}^K$, that exhibits both high quality and diversity together.

We begin by generating a large set of $M$ candidate outputs $ \{\x^{(i)}\}_{i=1}^M $ using i.i.d. sampling:
\begin{equation}
    \x^{(i)} = G_\theta(\z^{(i)}, \c), \quad \z^{(i)} \overset{\text{i.i.d.}}{\sim} p(\z).
\end{equation}

Let $\mathcal{I} = \{1, \ldots, M\}$ index the candidate samples. 
We associate each sample $i \in \mathcal{I}$ with a unary score $\u_i \in \mathbb{R}$ (e.g., CLIPScore~\cite{ramesh2022hierarchical} between image CLIP embedding and input caption text embedding) and each pair $(i, j)$ with a binary score $\b_{ij} \in \mathbb{R}$ (e.g., DINO~\cite{oquab2023dinov2} distances between two images). Concretely, 
\review{
\begin{align}
    \u_i &= f_{\text{CLIP}}(\x^{(i)}, \c) \label{eq:unary_v3} \\
    \b_{ij} &= 1 - \text{cosine}\left(f_{\text{DINO}}(\x^{(i)}), f_{\text{DINO}}(\x^{(j)}) \right) \label{eq:binary_v3}
\end{align}
where $f_{\text{CLIP}}$ computes the similarity between the input image and the target caption, and $f_{\text{DINO}}$ is the DINOv2 feature extractor.
Note that our method is general and accommodate many different choices of score functions as discussed later in Section~\ref{sec:different_diversity_scores}.
}

We introduce binary selection variables \( \y_i \in \{0, 1\} \) where \( \y_i = 1 \) indicates that candidate \( i \) is included in the next group. We define the group selection objective as:
\begin{align}
\label{eq:qip}
\max_{\y \in \{0,1\}^M} \quad 
& \sum_{i \in \mathcal{I}} \u_i \, \y_i + \lambda \sum_{\substack{i, j \in \mathcal{I} \\ i < j}} \b_{ij} \, \y_i \, \y_j \nonumber \\[4pt]
\text{subject to} \quad 
& \sum_{i \in \mathcal{I}} \y_i = K.
\end{align}

$\lambda$ is the hyperparameter that controls the relative weight between the unary and binary scores. The first term rewards individually strong outputs; the second promotes diversity by favoring dissimilar pairs. Solving this quadratic integer program (QIP) yields a subset of size \( K \) with desirable group properties. 
We use the branch-and-cut algorithm implemented by an off-the-shelf solver~\cite{gurobi} to solve the QIP.
Note that the formulation is model-agnostic and can accommodate any scoring functions, including functions that are not differentiable.

\subsection{Progressive Pruning for Efficient Selection}
\label{sec:progressive}
Naively applying group selection requires generating all $M$ candidates to completion, which is prohibitively expensive for recent compute-intensive models like Flux~\cite{flux}.
For example, generating $M=64$ samples over $T=20$ denoising steps requires $M \cdot T$ forward passes. 
Even on a modern GPU like NVIDIA H100, this results in a runtime of more than 3 minutes. 
To reduce this cost, we introduce a progressive filtering strategy that prunes candidates early using intermediate predictions.

\begin{figure*}[t!]
    \centering 
    \includegraphics[width=\linewidth]{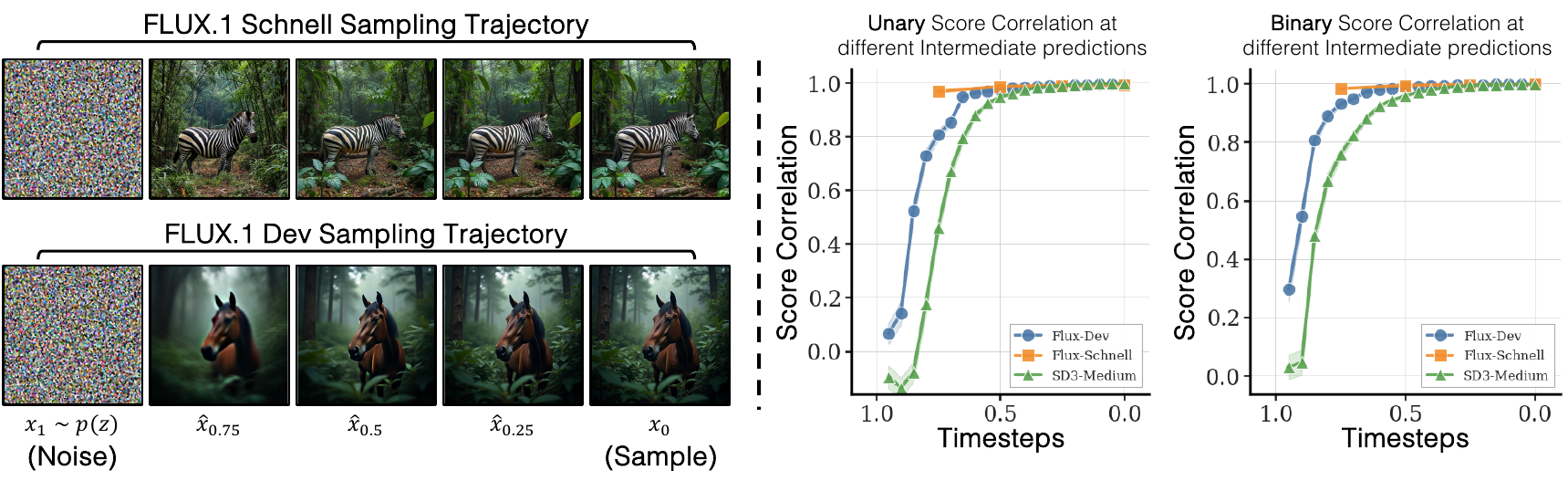}
    \caption{ 
    \textbf{Correlation Between Intermediate and Final Generation Scores.} 
    On the left, we show the reverse diffusion process, visualizing the intermediate predictions $\hat{\x_t}$ of the final image at different steps for FLUX.1 Schnell and FLUX.1 Dev models. We can observe that the intermediate predictions look similar to true final sample $\x_0$ for both the models.
    We further demonstrate this quantitatively by plotting the Spearman correlation of the Unary and Binary scores from $\hat{\x_t}$ versus final $\x_0$ scores, across different steps. For multistep models like FLUX.1 Dev and Stable Diffusion 3, the plots demonstrate strong correlations rapidly approaching 1.0, even at early timesteps. Note that for a timestep distilled model like Flux-Schnell, the correlation is high from the first denoising step. 
    This highlights the utility of using intermediate predictions for progressively filtering candidate samples.
    } 
    \label{fig:correlations}
    \vspace{-3mm}
\end{figure*}

\definecolor{codeblue}{rgb}{0.25,0.5,0.5}
\definecolor{codekw}{rgb}{0.85, 0.18, 0.50}
\definecolor{codesign}{RGB}{0, 0, 255}
\definecolor{codefunc}{rgb}{0.85, 0.18, 0.50}

\lstdefinelanguage{PythonFuncColor}{
  language=Python,
  keywordstyle=\color{blue}\bfseries,
  commentstyle=\color{codeblue},  %
  stringstyle=\color{orange},
  showstringspaces=false,
  basicstyle=\ttfamily\small,
  literate=
    {*} {{\color{codesign}*}} {1}
    {-} {{\color{codesign}-}} {1}
    {+} {{\color{codesign}+}} {1}
    {denoise} {{\color{codefunc}denoise}} {1}
    {unary_score} {{\color{codefunc}unary\_score}} {1}
    {binary_score} {{\color{codefunc}binary\_score}} {1}
    {SolveQIP} {{\color{codefunc}SolveQIP}} {1}
    {epsilon_theta} {{\color{codefunc}$\epsilon_\theta$}} {1} %
}

\lstset{
  language=PythonFuncColor,
  backgroundcolor=\color{white},
  basicstyle=\fontsize{9pt}{9.9pt}\ttfamily\selectfont,
  columns=fullflexible,
  breaklines=true,
  captionpos=b,
  mathescape=true, %
}

\begin{figure}[htbp] %
\begin{minipage}{0.8\linewidth}
\begin{algorithm}[H]
\caption{Efficient group inference}
\label{alg:main}
\begin{lstlisting}
# model: The diffusion model $\color{codeblue}\epsilon_\theta$
# zs: Initial noise vectors {z_i}
# N: Total number of denoising steps
# ts: The noise schedule {t_j}
# K: The target number of samples
# c: Conditioning information
# rho: The dropping ratio $\color{codeblue}\rho$ for pruning
def group_inference(model, zs, N, ts, K, c, rho):
    # Initialize the set of candidates from noise
    candidate_set = list(zs)

    # Denoising loop
    for j in reversed(range(1, N + 1)):
        intermediate_previews, next_latents = [], []
        
        # Get intermediate previews for all candidates
        for x_t in candidate_set:
            preview, x_next = denoise(x_t, ts[j], ts[j-1], c, model)
            previews.append(preview)
            next_latents.append(x_next)

        if len(candidate_set) > K:
            # Score previews and select the best subset
            u = unary_score(previews)
            b = binary_score(previews)
            
            # Prune candidates based on the dropping ratio rho
            m = max(K, int(len(candidate_set) *$~$(1 - $~$rho)))
            
            indices = SolveQIP(u, b, m)
            candidate_set = [next_latents[i] for i in indices]
        else:
            candidate_set = next_latents
            
    return candidate_set
\end{lstlisting}
\end{algorithm}
\end{minipage}
\end{figure}

\myparagraph{Intermediate pruning.}
We maintain a set $\mathcal{S}_t \subseteq \mathcal{I}$ of candidate indices at each step $t$.
For each sample in $\mathcal{S}_t$, we compute the intermediate prediction $\hat{\x}_{t}$, evaluate the unary and binary scores, and solve the QIP (Eq.~\ref{eq:qip}) to select the best subset. This subset becomes the next set $\mathcal{S}_{t-1}$, forming a nested sequence:
\[
\mathcal{S}_T \supset \mathcal{S}_{T-1} \supset \dots \supset \mathcal{S}_0.
\]
Once the set reaches the desired output group size $K$, we stop the pruning and complete the remaining denoising steps only for the selected samples.
See Algorithm \ref{alg:main} for the full procedure.

\myparagraph{Reliability of early predictions.}
In modern multi-step diffusion and flow-based models, the intermediate state $\x_t$ already encodes coarse information about the final generated sample $\x_0$. A common approximation of the final image at timestep $t$ is the predicted reconstruction:
\begin{equation}
\label{eq:x0hat}
    \hat{\x}_{t} = \x_t + t \cdot \epsilon_\theta(\x_t, t, \c),
\end{equation}
where $\epsilon_\theta$ predicts the noise or velocity at time $t$. 
Although these predictions are coarse, they are sufficient for computing the unary and binary scoring functions. 

To quantify this, we compute the correlation between the scoring functions (e.g., CLIP similarity or pairwise DINO diversity) evaluated on the intermediate images $\hat{\x}_{t}$ and the final output $\x_0$. Across a range of denoising steps, Figure~\ref{fig:correlations} (right) shows strong correlations (e.g., $r > 0.7$ after 5 steps for multi-step models, and $r > 0.95$ after the first step for distilled models), indicating that intermediate predictions are reliable proxies.  Figure~\ref{fig:correlations} (left) shows this visually. 
This high correlation enables us to safely rank candidates before they are fully denoised.

\subsection{Computational Complexity Analysis} 
To analyze the efficiency of progressive filtering, suppose we start with $M$ candidate samples and prune the set by a fixed ratio $\rho \in (0,1)$ at each denoising step until reaching a target set size $K$. Thus, the number of candidates at timestep $t$ is given by:
\begin{equation}
|\mathcal{S}_t| = \max\left(\rho^t M, K\right).
\end{equation}
Let $T$ denote the total number of denoising steps. We define the timestep $t^*$ at which the candidate set size first reaches or falls below the target $K$:
\begin{equation}
t^* = \left\lceil\frac{\log(K/M)}{\log(\rho)}\right\rceil.
\end{equation}
The total number of model evaluations \(f_\theta\) required throughout the process can be written as: 
\begin{equation}
M \cdot \frac{1 - \rho^{t^*}}{1 - \rho} + K \cdot (T - t^* + 1).
\end{equation}

In contrast, naive sampling without pruning would require \( M \cdot T \) model evaluations. For typical parameter settings (e.g., \(M=64\), \(K=4\), \(\rho=0.5\), \(T=20\)), our progressive filtering approach yields substantial compute savings (i.e., $184$ vs $1280$ evaluations, $\sim85\%$ reduction).
Our method has an overall complexity of $\mathcal{O}(M + K T)$.

\begin{figure*}[t!]
    \centering 
    \includegraphics[width=\linewidth]{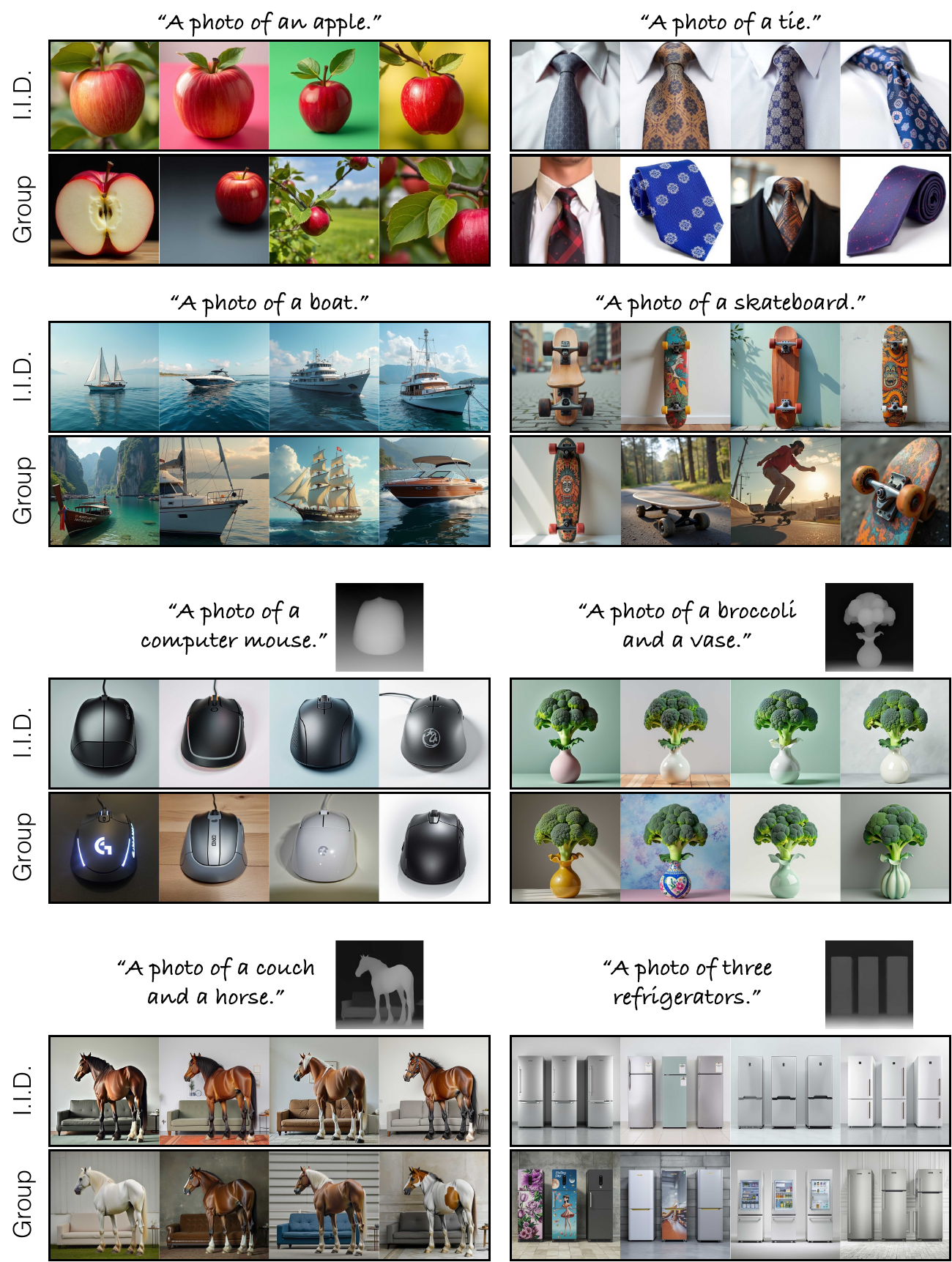}
    \caption{\textbf{Gallery of Results.} Qualitative results that show the advantage of our proposed group inference method over I.I.D. sampling for text-to-image generation and depth-to-image generation. Top row shows results with FLUX.1 Schnell, the second row uses FLUX.1 Dev, and the last two rows use FLUX.1 Depth as the base model.
    \review{
    For text-to-image generation, our method produces more diverse object poses and orientations, while for depth-to-image generation, it enhances color and texture diversity while adhering to the input depth condition.
    }} 
    \label{fig:qual_results_mp}
    \vspace{-5mm}
\end{figure*}

\section{Experiments} \label{sec:experiments}

We demonstrate the effectiveness of our proposed scalable group inference method across three different tasks: text-to-image generation, depth-conditioned generation, encoder-based image customization, and five different base models: FLUX.1 Schnell, FLUX.1 Dev, Stable Diffusion 3 (Medium), FLUX.1 Depth, and SynCD. 
The dataset and evaluation protocols used throughout all experiments are described next in \refsec{dataset}. 
Subsequently, in \refsec{baselines_tradeoff} and \refsec{inference_scaling}, we compare against prior methods along two axes: diversity-quality tradeoff, and inference-time scalability with different compute budget constraints. 
Finally, we present an ablation study to analyze the different components of our method, a \review{runtime analysis,} and their respective contributions. Please see the Appendix~\ref{sec:sup_results},~\ref{sec:sup_analysis},~\ref{sec:sup_details} for more results, analysis and ablations. 

\subsection{Dataset and Evaluation} \label{sec:dataset}
\myparagraph{Datasets.} We use the GenEval dataset~\cite{ghosh2023geneval}, validation split of the COCO 2017  dataset~\cite{lin2014microsoft}, and DreamBooth dataset~\cite{ruiz2022dreambooth} for text-to-image generation, depth-conditioned generation, and image customization, respectively. For depth-conditioned generation, we first extract the depth map using a recent method~\cite{depthanything}. 
Please refer to Appendix~\ref{sec:sup_details} for additional details.

\myparagraph{Models.}
We use several recent models, including FLUX.1 Dev~\cite{flux} and Stable Diffusion 3 Medium (SD3-M)~\cite{esser2024scaling}, which are flow-based models typically requiring 20-50 denoising steps. 
We also evaluate FLUX.1 Schnell, a timestep-distilled variant designed for efficient generation, typically using 1-8 steps. 
For depth-conditioned generation, we use FLUX.1 Depth, a model specifically trained for structural guidance based on depth maps. 
For customization, we use SynCD~\cite{kumari2025syncd}, a recent encoder-based image prompting model.
Unless otherwise specified, the sampling parameters for these models are fixed to the default values. A comprehensive list and inference parameters used can be found in Appendix~\ref{sec:sup_details_baselines}.

\myparagraph{Score Functions.} We use CLIP text-image similarity ~\cite{radford2021learning} to assess the quality of the individual samples (unary score) for the text-to-image and depth-to-image generation. For encoder-based image customization, we use cosine DINOv2~\cite{oquab2023dinov2} similarity between the input subject image and the output generated images for the unary score. Diversity (binary score) is computed for all tasks as one minus the cosine similarity between the DINOv2 patchwise features of all image pairs in the output set. 
\review{Our method can naturally accommodate a wide range of unary and binary scores, fitting the user's needs. Section~\ref{sec:different_diversity_scores} demonstrate this concretely.}

\review{
\myparagraph{User Study.}
We conduct two user preference studies to compare our method against each baseline on text-to-image generation. 
The first user study evaluates output diversity. In each comparison, the users are 
presented with two sets of 4 output images generated by two methods. The users are instructed to choose the set that has the higher variety. 
The second user study evaluates individual sample quality. For this study, the users are shown two images generated by two methods and asked to pick the one with higher quality. 
Both studies were conducted using Amazon Mechanical Turk (AMT) using prompts from our entire validation set. Each comparison was rated by three unique users, resulting in a total of 23,226 preference judgments.
}

\myparagraph{Runtime.} We measure inference runtime using wallclock time. Specifically, this is the time taken by each method to generate an output set of $K$ images (where $K=4$, unless specified otherwise) from a given input condition (i.e., a text prompt, depth map, or subject image). This measurement excludes initial model loading times and is averaged over 20 independent runs for each reported value. All runtime experiments utilize a single NVIDIA H100 GPU. Runtime comparisons based on the number of function evaluations (NFEs) are included in the appendix.

\myparagraph{Uncertainty Estimation.} We report standard errors for all quantitative results presented throughout our experiments. These standard errors are computed via bootstrapping with 1000 resamples. 

\begin{figure*}[t!]
    \centering 
    \includegraphics[width=1.0\linewidth]{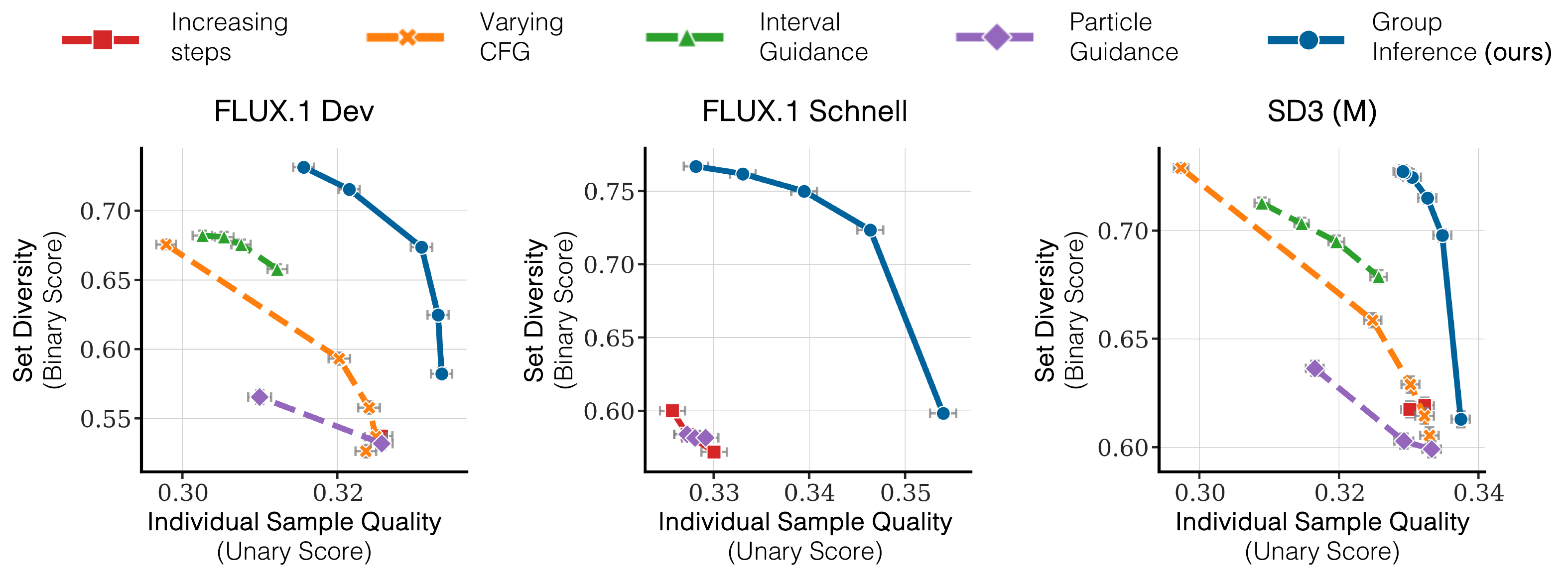}
    \vspace{-6mm}
    \caption{ \textbf{Quality and Diversity Pareto front for text-to-image models.} Each curve corresponds to a different inference strategy for three different text-to-image models (FLUX.1 Dev, FLUX.1 Schnell, and Stable Diffusion 3 Medium). Our proposed Group Inference ({\color{plt_blue} \textbf{blue}}) consistently dominates alternate methods (Increasing steps, Varying CFG, Interval Guidance, and Particle Guidance) achieving Pareto optimality and superior tradeoffs between quality and diversity across all methods. Varying CFG and Interval Guidance do not apply to the distilled model (FLUX.1 Schnell). 
    }
    \label{fig:baselines_UB_t2i}
    \vspace{-3mm}
\end{figure*}

\begin{table*}[ht!]
\centering
\resizebox{1.0\linewidth}{!}{

\begin{tabular}{llcccc}
\toprule
& & \multicolumn{2}{c}{\textbf{Diversity}} & \multicolumn{2}{c}{\textbf{Quality}} \\
\cmidrule(lr){3-4} \cmidrule(lr){5-6}
\textbf{Model} & \textbf{Comparison} & \textbf{Ours pref.} & \textbf{Baseline pref.} & \textbf{Ours pref.} & \textbf{Baseline pref.} \\
\midrule
\multirow{4}{*}{FLUX.1 Dev} & Ours vs Low-CFG & \textbf{88.3\%} & 11.70\% & \textbf{85.6\%} & 14.4\% \\
& Ours vs Interval Guidance & \textbf{53.4\%} & 46.6\% & \textbf{58.4\%} & 41.6\% \\
& Ours vs Particle Guidance & \textbf{81.2\%} & 18.8\% & \textbf{79.4\%} & 20.6\% \\
\midrule
\multirow{4}{*}{FLUX.1 Schnell} & Ours vs Low-CFG &
\multicolumn{4}{c}{\textbf{N/A}} \\
& Ours vs Interval Guidance & 
\multicolumn{4}{c}{\textbf{N/A}} \\
& Ours vs Particle Guidance & \textbf{55.5\%} & 44.5\% & \textbf{62.3\%} & 37.7\% \\
\midrule
\multirow{4}{*}{SD3 (M)} & Ours vs Low-CFG & \textbf{76.8\%} & 23.2\% & \textbf{80.8\%} & 19.20\% \\
& Ours vs Interval Guidance & \textbf{58.1\%} & 41.9\% & \textbf{57.9\%} & 42.10\%\\
& Ours vs Particle Guidance & 
\textbf{78.9\%} & 21.1\% & \textbf{85.9\%} & 14.1\%\\
\bottomrule
\end{tabular}
}

\caption{\review{\textbf{User preference comparison between our method and baselines.} Results from our user study demonstrate that our method is consistently preferred over alternative inference strategies. Across three different text-to-image models (FLUX.1 Dev, FLUX.1 Schnell, and Stable Diffusion 3 Medium), users consistently chose our generations for both diversity and quality. Note that comparisons against Low-CFG and Interval Guidance are not applicable (N/A) for the distilled FLUX.1 Schnell model.}}
\label{tab:user_preference}
\end{table*}

\begin{figure}[t!]
    \centering 
    \includegraphics[width=1.0\linewidth]{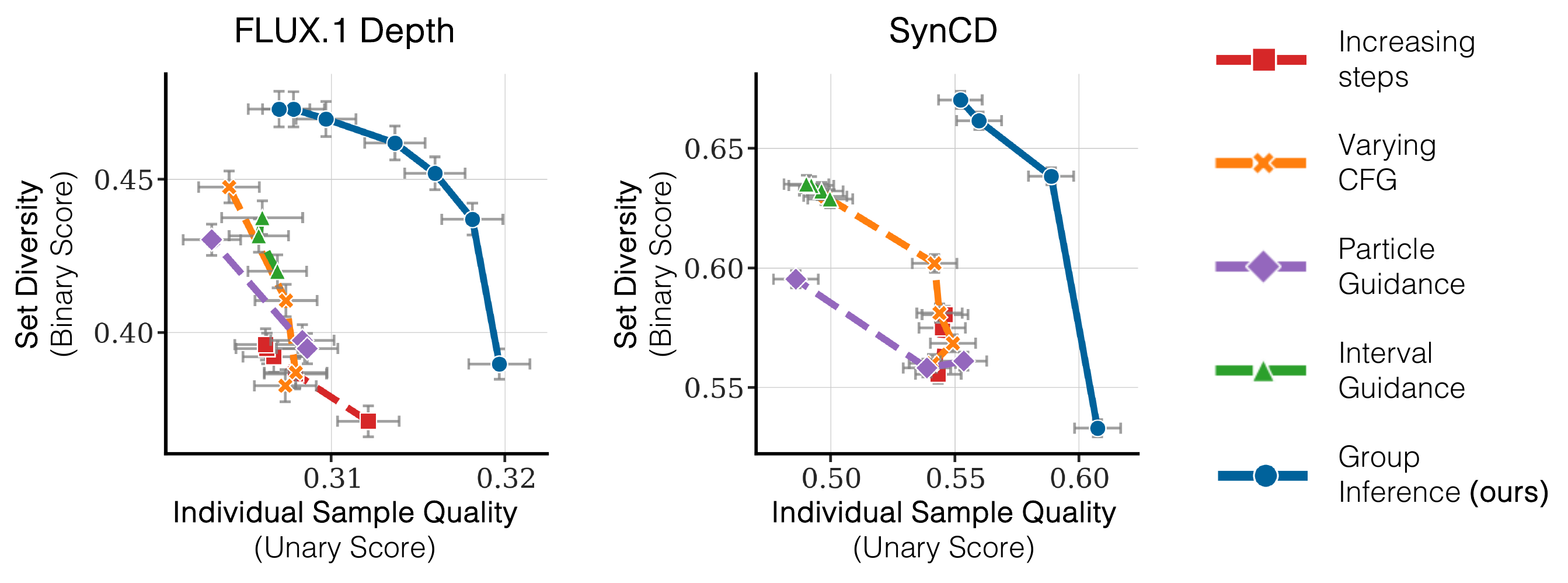}
    \vspace{-6mm}
    \caption{\textbf{Quality and Diversity Pareto front for additional tasks.} Each curve corresponds to a different inference strategy for depth conditioned generation (left, FLUX.1 Depth) and image prompting (right, SynCD). Our proposed Group Inference ({\color{plt_blue} \textbf{blue}}) consistently dominates alternative methods—Increasing steps, Varying CFG, Interval Guidance, and Particle Guidance—achieving Pareto optimality and superior tradeoffs between quality and diversity across all methods. 
    } 
    \label{fig:baselines_UB_other_tasks}
\end{figure}

\begin{figure}[t!]
    \centering 
    \includegraphics[width=\linewidth]{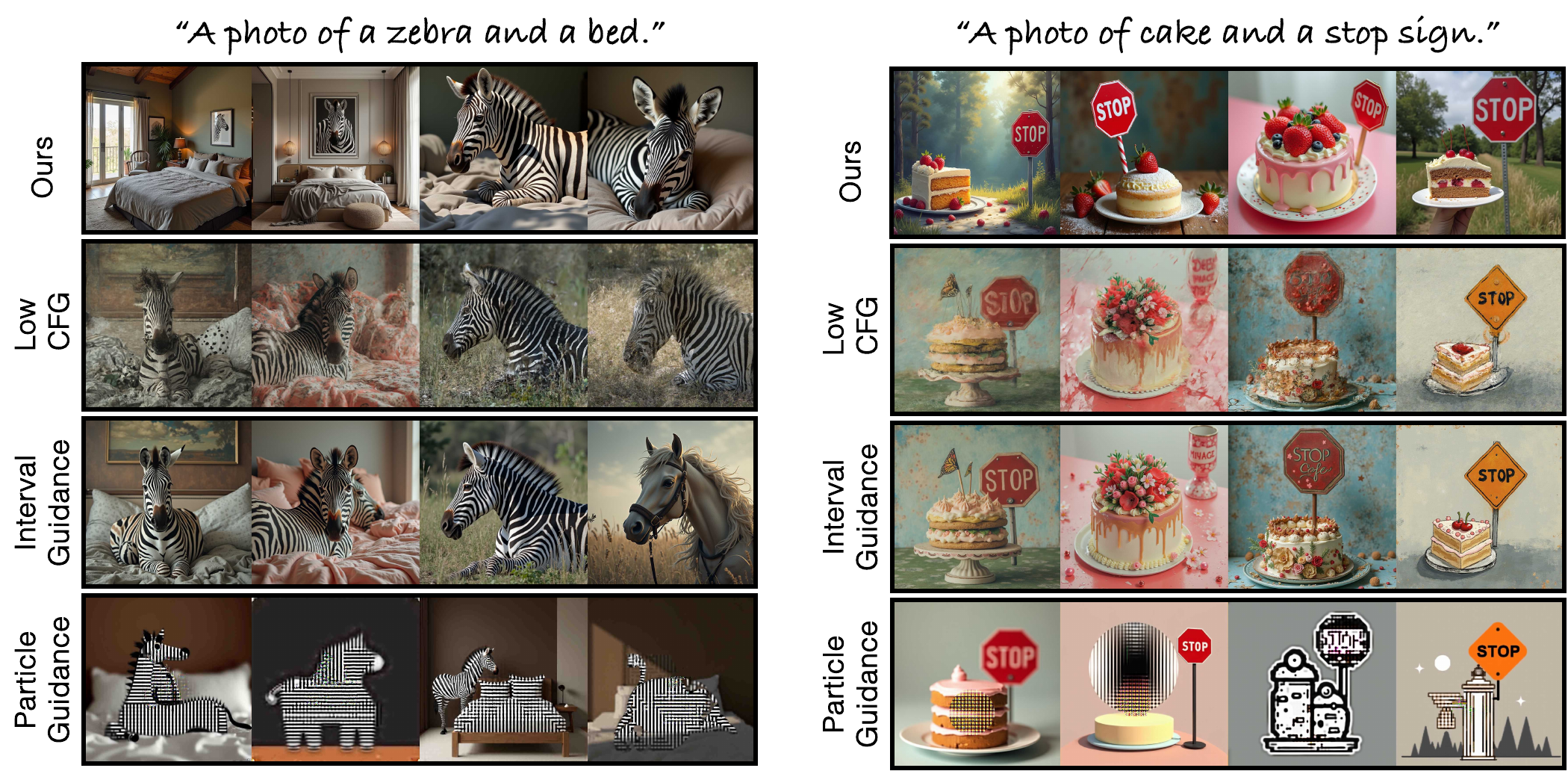}
    \caption{ \textbf{Qualitative results.} 
    We compare our proposed method (top row) against alternative inference strategies targeting an improved Quality-Diversity tradeoff with FLUX.1 Dev base model. To ensure a fair comparison, baseline methods were configured to approximate the diversity level achieved by our approach. The precise parameters of each baseline, and a comparison at other configurations is shown in the supplement. The result demonstrates that: (i) employing a low Classifier-Free Guidance (CFG) scale to increase diversity results in diminished image quality; (ii) Interval Guidance exhibits reduced adherence to the input text prompt; and (iii) Particle Guidance, by actively altering sampling trajectories, tends to produce less natural images. In contrast, our method outputs a set of diverse outputs while maintaining good image quality and prompt fidelity.
    } 
    \label{fig:visual_comparison}
\end{figure}

\begin{figure*}[t!]
    \centering 
    \includegraphics[width=0.95\linewidth]{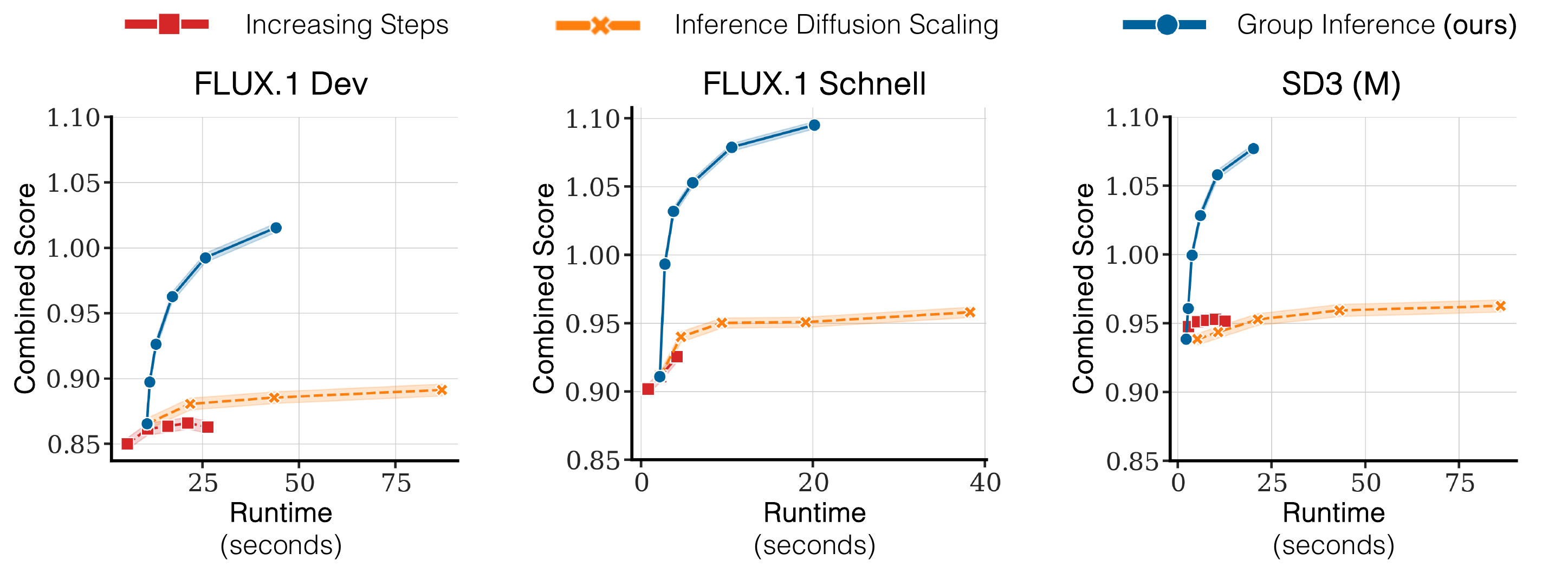}
    \caption{\textbf{Performance at different runtimes.}
    The Increasing Steps ({\color{plt_red}\textbf{red}}) baseline shows limited gains with additional computation. The Inference Diffusion Scaling~\cite{ma2025inference} method ({\color{plt_orange}\textbf{orange}}), which increases sample count through independent I.I.D. generations, requires substantially more runtime for marginal improvements. In contrast, our proposed Group Inference ({\color{plt_blue}\textbf{blue}}) achieves significantly better performance–runtime tradeoffs, quickly outperforming both baselines with minimal overhead. Observation holds across models.
    } 
    \vspace{-13pt}
    \label{fig:inference_scaling}
\end{figure*}

\begin{figure*}[t!]
    \centering 
    \includegraphics[width=1.0\linewidth]{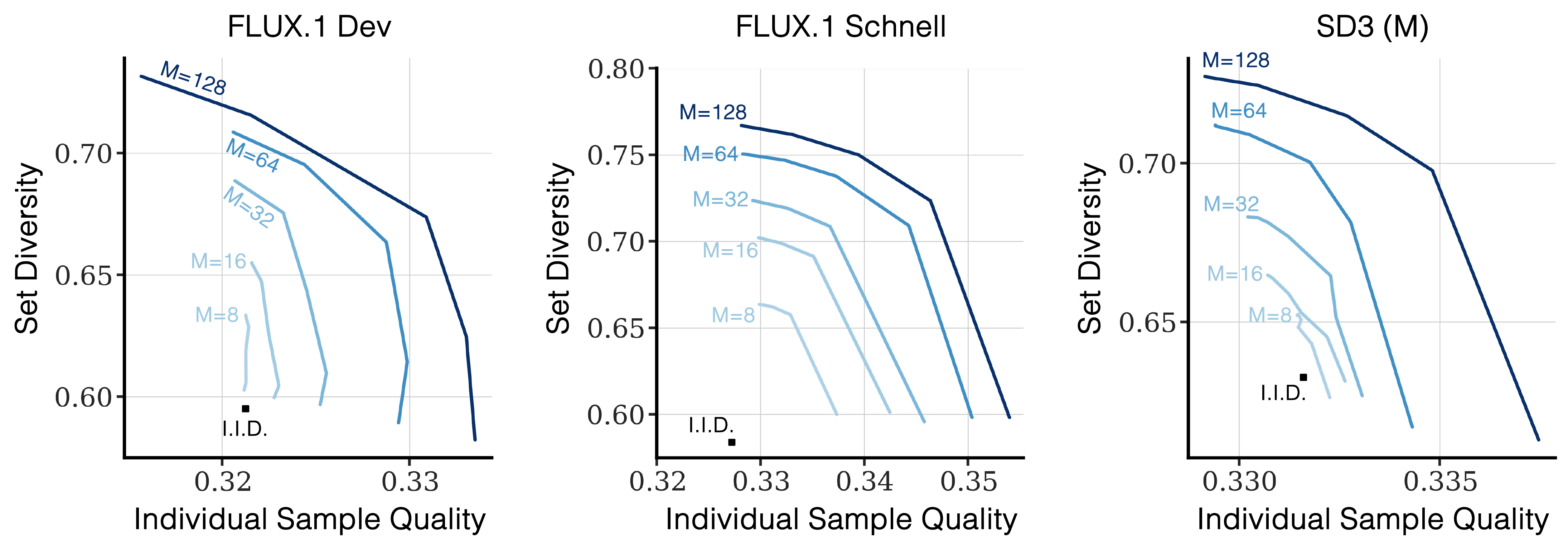}
    \vspace{-5mm}
    \caption{
    \review{\textbf{Improvements as the number of initial samples M is increased.} We show how the sample quality (measured with CLIP) and the set diversity (measured with DINO) improves as the number of initial starting samples is increased from 4 to 128.
    }}
    \vspace{-13pt}

    \label{fig:m_scaling}
\end{figure*}

\subsection{Baselines and the Diversity-Quality Tradeoff}
\label{sec:baselines_tradeoff}
In generative modeling, a fundamental tradeoff often exists between optimizing for the perceptual quality of individual samples and ensuring a diverse set of outputs~\cite{zhu2017toward,ho2021classifier,brock2018large}. 
Many prior methods implicitly or explicitly navigate this spectrum. In this subsection, we compare our proposed approach to existing approaches proposed to achieve a more favorable Pareto frontier in the diversity-quality space.
Figure~\ref{fig:baselines_UB_t2i} plots the quality and diversity Pareto front for text-to-image models (FLUX.1 Dev, FLUX.1 Schnell, and SD3 (M)). Table~\ref{tab:user_preference} shows the results of a pairwise user preference study between our method and the baselines. Figure~\ref{fig:baselines_UB_other_tasks} plots the quality diversity tradeoff for additional models.
Qualitative comparisons are shown in Figure~\ref{fig:visual_comparison}.

Across all baselines, our proposed method consistently achieves a superior diversity-quality tradeoff. As illustrated by the \textcolor{plt_blue}{\textbf{blue}} line in Figures~\ref{fig:baselines_UB_t2i} and ~\ref{fig:baselines_UB_other_tasks}, our approach dominates the Pareto fronts of all evaluated baselines, yielding better diversity for a given level of quality, or higher quality for a comparable level of diversity. Comparisons with additional metrics are shown in the Appendix.

\myparagraph{Increasing Denoising Steps.} We first consider the impact of simply increasing the number of denoising steps during sampling. While more steps can sometimes refine details, we find this has a minimal effect on meaningfully shifting the diversity-quality balance for the models under study, as shown by the \textcolor{plt_red}{\textbf{red}} line in Figures~\ref{fig:baselines_UB_t2i} and~\ref{fig:baselines_UB_other_tasks}. 

\myparagraph{Varying CFG.} Next, we examine the widely used technique of varying the Classifier-Free Guidance scale ~\cite{ho2021classifier} (CFG).
As depicted by the \textcolor{plt_orange}{\textbf{orange}} line in Figures~\ref{fig:baselines_UB_t2i} and~\ref{fig:baselines_UB_other_tasks}, systematically altering the CFG scale traces a distinct tradeoff curve.
Notably, low CFG values (e.g., CFG=1) largely increase output diversity but often at the cost of a sharp degradation in sample quality and prompt alignment. This is visually seen through the poor image quality in the second row in Figure~\ref{fig:visual_comparison} where a low CFG value is used. 
The results of the user study in Table~\ref{tab:user_preference} further corroborate these observations. 

\myparagraph{Interval Guidance.} Interval guidance~\cite{kynkaanniemi2024applying} attempts to refine this by applying CFG selectively only during a subset of the denoising timesteps. We conduct a sweep across various interval configurations, with the results shown as the \textcolor{plt_green}{\textbf{green}} line in Figures~\ref{fig:baselines_UB_t2i} and~\ref{fig:baselines_UB_other_tasks}.
Consistent with the original findings for Interval Guidance and third row in Figure~\ref{fig:visual_comparison}, this approach can offer image quality improvements over a standard CFG sweep. However, it still performs worse than our method in terms of both quality and diversity.
\review{For instance, in the example of zebra and bed on the left in Figure~\ref{fig:visual_comparison}, Interval Guidance has reduced diversity of zebra poses and does not generate a bed for two of the four outputs. Similarly, in the right example, interval guidance does not always generate a stop sign.}
Moreover, both standard CFG sweeping and Interval Guidance do not apply to distilled models like FLUX.1 Schnell, as these models do not use guidance mechanisms.

\myparagraph{Particle Guidance.}
We also evaluate Particle Guidance~\cite{corso2023particle}, which optimizes a binary potential function to encourage diversity during inference. Following the original work, we use DINO features for the diversity term. 
As illustrated by the \textcolor{plt_purple}{\textbf{purple}} line, Particle Guidance can indeed increase output diversity. However, this comes with a sharp decrease in individual sample quality (fourth row in Figure~\ref{fig:visual_comparison}), as direct optimization of the binary potential actively alters the sampling trajectory. This can push the output samples off the learned data manifold, leading to less natural and artifact-prone images. Furthermore, Particle Guidance requires a substantial memory cost due to the necessity of computing gradients and backpropagating through the binary potential. 
This reliance on gradient computation also makes the method unsuitable for non-differentiable potential functions.

\subsection{Inference Scaling Analysis}
\label{sec:inference_scaling}
Scaling computational resources at the test time to enhance model performance is an increasingly useful paradigm in machine learning.
For diffusion models, a native mechanism for test-time scaling involves increasing the number of denoising steps. Although this can initially lead to improved sample quality, this approach often yields diminishing returns; beyond a certain point, additional denoising steps provide progressively smaller gains in quality. 
In Figure~\ref{fig:inference_scaling}, we illustrate the impact of various test-time scaling strategies on the group objective (defined in \refeq{qip}), evaluated across various computational budgets.

Our first baseline (Figure~\ref{fig:inference_scaling}, \textcolor{plt_red}{\textbf{red}} line) allocates increased compute to a greater number of denoising steps for a fixed number of initial samples $M$. Consistent with existing findings~\cite{ma2025inference}, this approach demonstrates minimal improvement in our combined score, with the curve quickly plateauing. %
Inference Diffusion Scaling~\cite{ma2025inference}, proposes an alternative that utilizes the additional compute budget to perform a search over multiple random seeds. For a fair comparison, we implement this baseline with the CLIP text-image similarity as the verifier. 
This method does not incorporate intermediate predictions and does not consider any pairwise terms. Consequently, it is not effective in improving the group objective, as shown through the \textcolor{plt_orange}{\textbf{orange}} line. 

In contrast, our method invests in the inference budget to increase the number of initial samples. As depicted by the \textcolor{plt_blue}{\textbf{blue}} line in Figure~\ref{fig:inference_scaling}, this approach produces consistent improvements in the combined group score.
\review{Figure~\ref{fig:m_scaling} further shows the improvement in both the quality and diversity of the outputs as the number of initial samples is gradually increased from 4 to 128 across three different base models.}

\begin{figure}[t!]
    \centering 
    \includegraphics[width=1.0\linewidth]{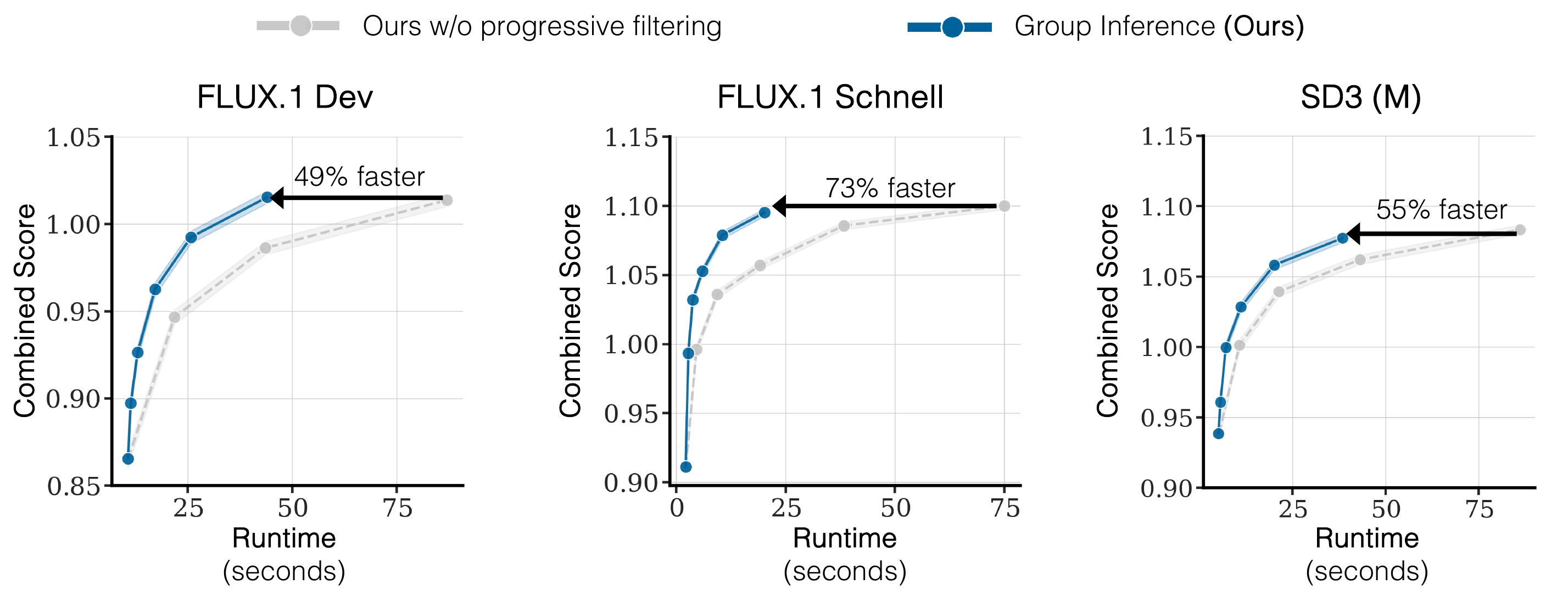}
    \vspace{-5mm}
    \caption{ \textbf{Importance of progressive pruning.} Across multiple base generative models, progressive pruning consistently enables our method to select candidates efficiently and shows substantial speedups-49\%, 73\%, and 55\% faster for comparable combined group scores.} 
    \label{fig:abl_progressive_filtering}
\end{figure}

\myparagraph{Ablating progressive filtering.}
The importance of progressive filtering is shown in Figure~\ref{fig:abl_progressive_filtering}. We compare our complete method, which utilizes progressive filtering with intermediate predictions $\hat{\x}_{t}$ (\textcolor{plt_red}{\textbf{red}} line), against a variant that performs full denoising for all $M$ candidate samples  without such filtering (\textcolor{gray}{\textbf{gray}} line). 
Demonstrating its effectiveness across different architectures, our approach achieved comparable group scores while requiring up to 73\% less runtime.
Please see the appendix for additional ablation studies. 

\begin{figure}[t!]
    \centering 
    \includegraphics[width=\linewidth]{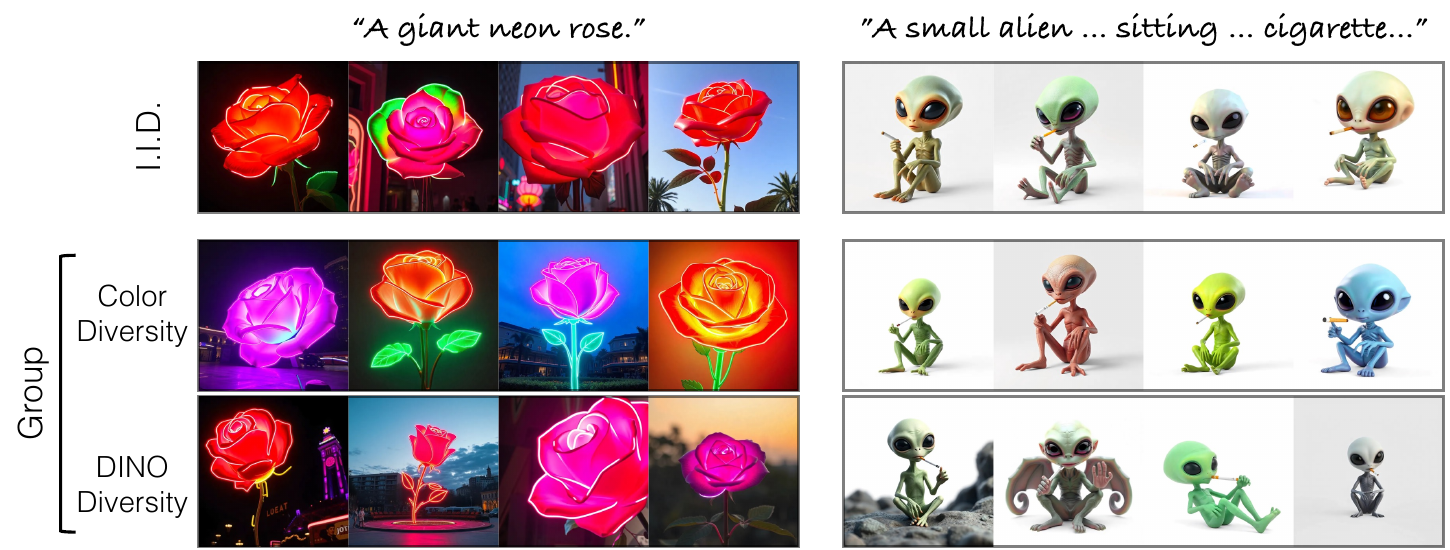}
    
    \caption{ 
    \review{
    \textbf{Accommodating different pairwise objectives.} 
    Compared to baseline I.I.D. sampling (top row), our method allows for targeted diversity by defining different pairwise objectives. 
    The second and third rows show results where the unary quality term is identical but the pairwise binary term is varied. The middle row uses a color-based binary term, while the bottom row uses a DINO-based binary term to achieve semantic and structural diversity.
    }
    }
    \label{fig:different_binary}
\end{figure}

\review{
\subsection{Different Diversity Objectives}
\label{sec:different_diversity_scores}

Next, we show that our approach is general and can accommodate different pairwise binary objectives by simply swapping the binary term in our quadratic integer programming objective, as demonstrated in Figure~\ref{fig:different_binary}.
Let us consider the example shown on the left corresponding to the caption ``a giant neon rose.''
Standard I.I.D. sampling (top row) produces a set of visually redundant images. All four roses are red and share a similar pose.

In contrast, the bottom two rows are generated using our method with an identical unary quality term (CLIP text-image similarity) but different binary diversity objectives. The middle row uses a direct color based dissimilarity as the binary term. This successfully steers the outputs towards a set of images with varied and distinct color schemes. For the rose example on the left, this results in a vibrant set that includes blue, orange, and pink neon variants.
In the bottom row, we use a DINO diversity metric that captures more semantic features when comparing the pairwise distances. This change directs the model to produce a set with higher structural variance. As seen with the rose example, this yields outputs with different poses and camera angles. 

This direct comparison underscores a key strength of our approach: the ability to seamlessly integrate different notions of diversity to achieve targeted, user-defined visual outcomes.
}

\section{Discussion, Broader Impacts, and Limitations}
\label{sec:conclusion}
In this paper, we have introduced scalable group inference, a novel method to generate diverse, high-quality sets of samples by formulating the selection as a quadratic integer program and leveraging intermediate predictions for improving the runtime efficiency. Our efficient approach significantly enhances group diversity and quality compared to existing baselines across various generative tasks. Still, our method has several limitations. 

\begin{figure}[t!]
    \centering 
    \includegraphics[width=\linewidth]{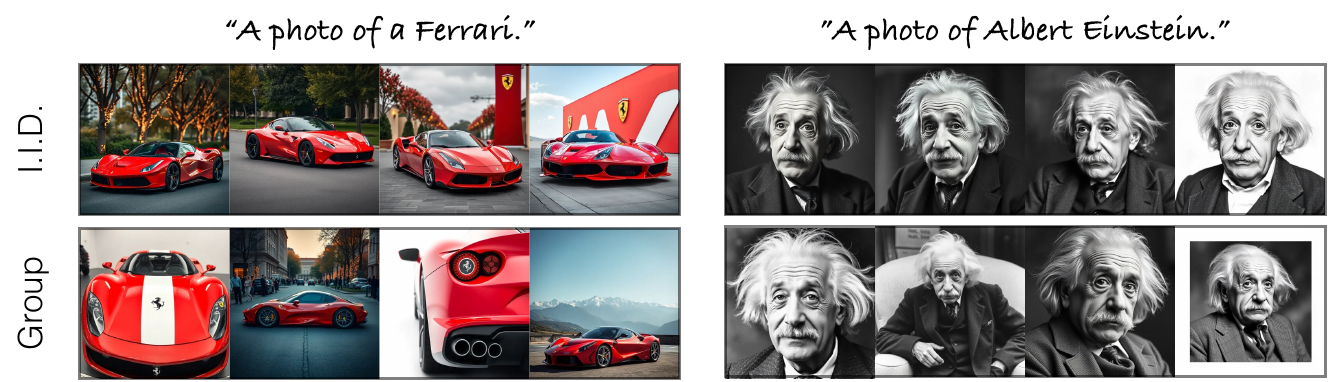}
    
    \caption{ 
    \review{
    \textbf{Failure cases.} 
    The performance of our method depends on the diversity of the initial candidate pool. 
    (Left) For the prompt ``A photo of a Ferrari,'' the base model (FLUX.1 Schnell) exhibits a strong color bias, exclusively generating red cars. Consequently, our method can find varied poses but is unable to produce a color-diverse set. (Right) Similarly, for ``a photo of Albert Einstein,'' the base model only generates black-and-white images, constraining our method from finding any color photographs.
    }
    }
    \label{fig:failure_cases}
\end{figure}

First, our method relies on the base generative model's ability to produce a sufficiently diverse and high-quality initial candidate pool. Consequently, if the underlying model generates outputs of inherently poor quality or suffers from significant mode collapse, the efficacy of Scalable Group Inference in identifying an optimal set will be inherently constrained, as our method selects from, rather than intrinsically enhances, these initial candidates. This is visually illustrated in Figure~\ref{fig:failure_cases}.

\review{Second, our method assumes that the unary (quality) and binary (diversity) scores are fast to compute. If evaluating these scores, especially the pairwise diversity metric across a large candidate set, is computationally intensive, the runtime benefits of our scalable optimization would be reduced. }

Nevertheless, our method offers a path to more user-centric systems that efficiently output diverse, high-quality sets of options, and enhance creative exploration. This capability can significantly reduce the iterative burden in content generation across various domains. %
Concurrently, the increased efficiency in generating diverse sets of synthetic media could also have potential for misuse, such as creating more varied and potentially harder-to-detect misleading content, demanding proactive ethical guidelines and mitigation strategies.

\begin{ack}
We thank Daniel Cohen-Or and Sheng-Yu Wang for their helpful comments and discussion. We are also grateful to Nupur Kumari for proofreading the draft. The project is partially supported by Snap Research, NSF IIS-2239076, DARPA ECOLE, and
the Packard Fellowship.

\end{ack}

\small
\bibliographystyle{unsrt}
\bibliography{main}

\appendix
\clearpage

Section~\ref{sec:sup_results} presents additional qualitative and quantitative results obtained by our method across multiple different base generative models and tasks. 
Section~\ref{sec:sup_analysis} provides additional analysis of the different components of our method. 
Section~\ref{sec:sup_details} details the datasets and the implementation settings used for each of the baseline methods. 

\section{Analysis} \label{sec:sup_analysis}
In this section, we provide additional analysis of the different components of our method.

\begin{figure}[t!]
    \centering 
    \includegraphics[width=\linewidth]{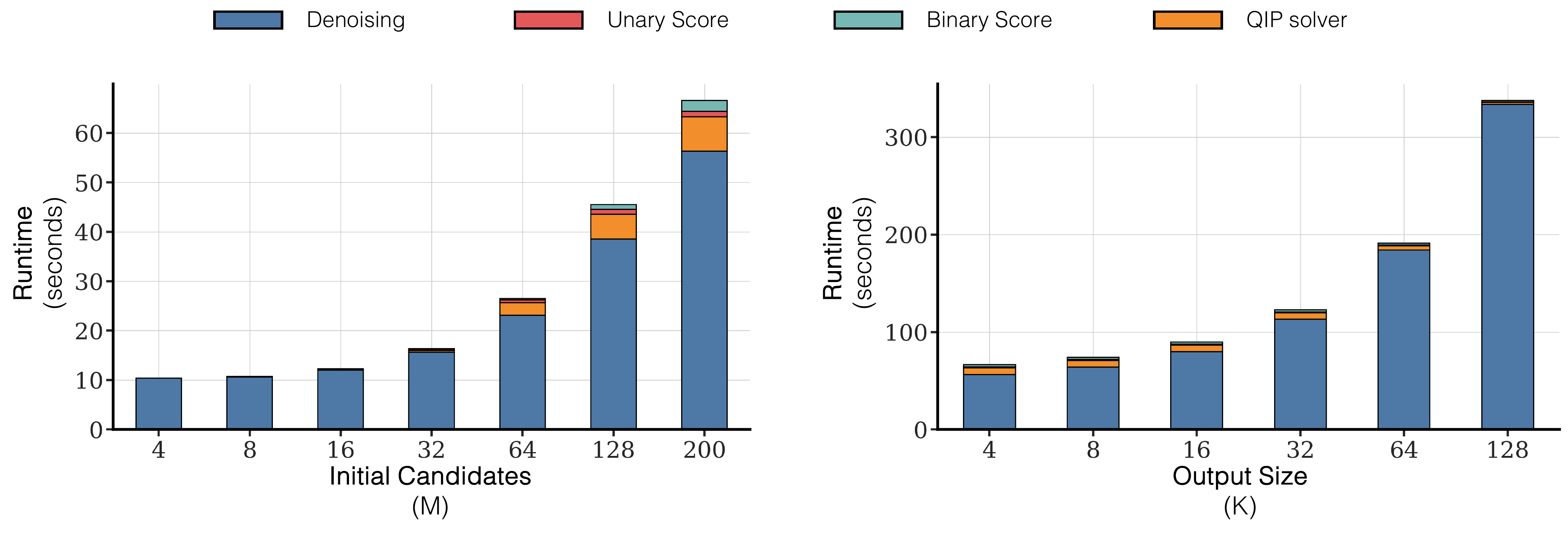}
    \caption{ 
    \textbf{Runtime breakdown.} We show a runtime breakdown of our method using FLUX.1 Dev model as the number of initial candidates (M, left) and the output set size (K, right) is increased. On the left plot, the output size is fixed to 4 and in the right plot the initial candidate size is fixed to 200. Across all settings, the runtime is dominated by the denoising step.
    } 
    \label{fig:sup_runtime_breakdown}
\end{figure}

\myparagraph{Runtime breakdown.}
In Figure~\ref{fig:sup_runtime_breakdown} we show the runtime breakdown of different steps in the pipeline for the FLUX.1 Dev base model using CLIP text image similarity as the unary score and DINOv2 diversity as the binary score. 
On the left, we fix the output set size K to be 4 and increase the initial candidate size from 4 to 200.
On the right, we fix the initial candidate size to be 200, and increase the output set size from 4 to 128. 
Note that across all settings, the runtime cost incurred by the QIP solver and the score computation is negligible compared to the forward pass of the denoising transformer. 

\begin{figure*}[t!]
    \centering 
    \includegraphics[width=\linewidth]{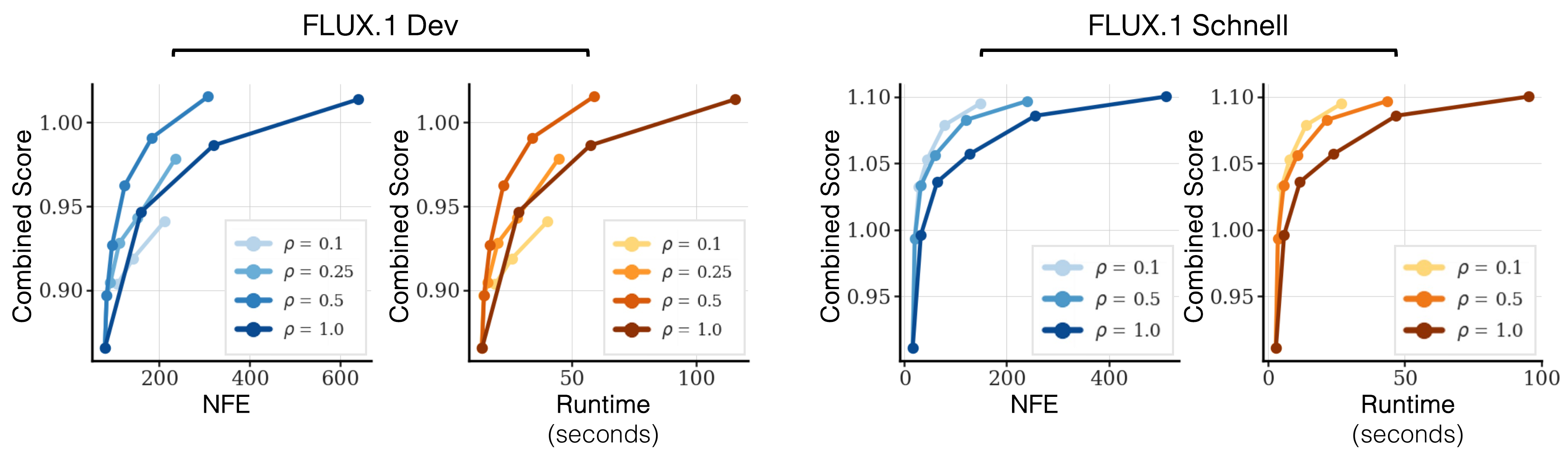}
    \caption{ 
    \textbf{Effects of different dropping ratio $\rho$.} We show the effects of different dropping ratios $\rho$ for two different base models: FLUX.1 Dev and FLUX.1 Schnell.  
    } 
    \label{fig:sup_ablation_rho}
\end{figure*}

\myparagraph{Different $\rho$ values.}
Figure~\ref{fig:sup_ablation_rho} illustrates the effects of varying pruning ratios ($\rho$) on the FLUX.1 Dev and FLUX.1 Schnell models.
The figure presents both the Number of Function Evaluations (NFE) (left plot) and the wallclock runtime on a single NVIDIA H100 (right plot).
Across all plots, a pruning ratio of $\rho=1.0$ signifies no progressive pruning.
For the FLUX.1 Dev model, lower pruning ratios (e.g., $\rho=0.1$ and $\rho=0.25$) are overly aggressive, leading to suboptimal scores.
Conversely, a pruning ratio of $\rho=1.0$ (no candidate filtering) achieves a good combined score but incurs a high inference cost.
A pruning ratio of $\rho=0.5$ strikes an effective balance, yielding higher scores without excessive computational cost. We use $\rho=0.5$ for all FLUX.1 Dev experiments. 

A different trend observed for distilled FLUX.1 Schnell model. It can accommodate a more aggressive pruning ratio, such as $\rho=0.1$, without a noticeable decrease in the score. 
This can be attributed to the better reliability of the intermediate predictions for the distilled models, as shown in Figure 3 of the main paper.

\myparagraph{Efficient decoding.}
Our method uses an efficient decoder ~\cite{bohan2023tiny} to decode all intermediate predictions for progressive pruning. In Figure~\ref{fig:sup_ablation_vae} we ablate the use of efficient decoder and show that across both FLUX.1 Dev and FLUX.1 Schnell, using an efficient decoder improves the runtime without sacrificing the score.

\myparagraph{Evaluation with different score functions.} Figure~\ref{fig:baselines_UB_t2i} in the main paper shows the quality and
diversity Pareto front for the text to image generation task. That figure uses CLIP text-image similarity (Equation~\ref{eq:unary_v3}) as the quality score and DINO diversity~\ref{eq:binary_v3} as the diversity score. Next, we evaluate our method using several additional score functions that are not used by our method for selection in Figure~\ref{fig:baselines_UB_other_metrics}. 
The top row uses Image Reward~\cite{xu2023imagereward} for measuring quality of samples and depth features to measure diversity. Image Reward is a network that is trained to learn human preferences for text-to-image generation. The depth diversity is calculated with the DepthAnything V2 model~\cite{yang2024depth}. The bottom row uses BLIP2~\cite{li2023blip} to measure the quality and CLIP features to compute the diversity. Figure~\ref{fig:baselines_UB_other_metrics} shows a comparison with three different base models: FLUX.1 Dev (left), FLUX.1 Schnell (middle), and Stable Diffusion 3 medium (right). Across each model, our proposed group inference shows a better trade off between quality and diversity. Note that particle guidance obtains slightly better BLIP2 score than our method for Stable Diffusion 3 (M). However, the outputs generated by particle guidance have artifacts. This is also reflected by a low score for other metrics (Image Reward and CLIP), and a worse user preference score.

\section{Additional Results} \label{sec:sup_results}

\myparagraph{Qualitative results.}
In Figures~\ref{fig:supp_results_depth_a},~\ref{fig:supp_results_depth_b}, and~\ref{fig:supp_results_syncd_a} we show additional visual examples of our method. 
Across multiple models and tasks, our method consistently outputs samples that are more diverse, and without any degradation in the quality. 
Similar to the Figure~\ref{fig:visual_comparison} in the main paper, Figure~\ref{fig:supp_visual_comparison} shows additional visual comparison to baselines. In these figures, the Low CFG baseline uses a CFG value of 1.0. Interval Guidance uses an interval of $[0.6, 0.4]$, and particle guidance uses a coefficient value of 100. 

\myparagraph{Correlation analysis.}
Figure~\ref{fig:correlations} in the main paper shows the correlation between the scores computed with the final image and the intermediate images. In Figure~\ref{fig:sup_correlations_all}, we show that a similar correlation trend is visible in other base models (FLUX.1 Depth and SynCD). FLUX.1 Depth and SynCD show the CLIP text image similarity as the unary scores, and DINO diversity as the binary scores. This is consistent with our observations in the main paper.

\section{Implementation Details} \label{sec:sup_details}
Section~\ref{sec:sup_details_baselines} first provides implementation details and hyperparameters used for all settings shown in Figures 4, 5, and 6 of the main paper.  
Section~\ref{sec:sup_details_datasets} lists details about the datasets used for each task. 

\subsection{Baselines} \label{sec:sup_details_baselines}

\myparagraph{Increasing steps.} 
For FLUX.1 Dev, Stable Diffusion 3 Medium, FLUX.1 Depth, and SynCD, we consider the timesteps 10, 20, 30, 40, and 50. 
For the distilled model, FLUX.1 Schnell, we consider the timesteps 1, 2, 4, and 8.

\myparagraph{Varying CFG.} 
For FLUX.1 Dev, FLUX.1 Depth, and SynCD we consider the CFG values 1, 2, 3, 4, and 5. 
For Stable Diffusion 3 Medium, we consider the CFG values 1, 5, 10, and 15. 
Note that FLUX.1 Schnell does not use CFG.

\myparagraph{Interval guidance.} 
For FLUX.1 Dev, FLUX.1 Depth, Stable Diffusion 3 Medium, and SynCD we consider the guidance intervals $[0.9, 0.1]$, $[0.8, 0.2]$, $[0.7, 0.3]$, and $[0.6, 0.4]$. 
Note that FLUX.1 Schnell does not use CFG, and this baseline is not applicable.

\myparagraph{Particle guidance.} 
For the Particle Guidance baseline, we consider coefficient values 0, 10, 50, 100, and 200. Note that this baseline significantly increases the memory consumption during inference. 

\myparagraph{Inference diffusion scaling~\cite{ma2025inference}.} Figure~\ref{fig:inference_scaling} of the main paper shows a comparison to Inference Diffusion Scaling~\cite{ma2025inference}, a concurrent work, that shows an improvement in the quality of samples. We follow the results in their paper and use random search as the strategy. For a fair comparison, we use the same CLIP text-image-similarity as the verifier. 

\myparagraph{Group inference (ours).} 
In Figures~\ref{fig:baselines_UB_t2i},~\ref{fig:baselines_UB_other_metrics}, and~\ref{fig:baselines_UB_other_tasks} of the main paper, we vary the $\lambda$ defined in Equation~\ref{eq:qip} while keeping the input samples $M$ fixed. We use $M=128$ for FLUX.1 Dev, FLUX.1 Schnell, SD3 (M), and SynCD.
Note that varying the weighting factor $\lambda$ does not change the runtime, and only shows the trade-off between the diversity of samples in the generated output set and the individual quality. For FLUX.1 Dev, FLUX.1 Depth, SynCD we use $\rho=0.5$. For SD3 (M) we use a higher $\rho=0.75$, and for timestep distilled model FLUX.1 Schnell $\rho=0.1$ in all experiments.  

In Figures~\ref{fig:inference_scaling},~\ref{fig:m_scaling} and~\ref{fig:abl_progressive_filtering} of the main paper, we want to study the performance at different runtimes, and therefore we fix the weighting factor $\lambda=1$ but vary the number of input samples $M$ from 4 to 128.

\myparagraph{Choice of scores.} 
Unless specified otherwise, FLUX.1 Dev, FLUX.1 Schnell, SD3 (M), Flux.1 Depth use CLIP text-image similarity as the unary score, and DINO diversity as the binary score. SynCD uses DINO target image similarity as the unary score, and DINO diversity as the binary score across all results.
\review{Figure~\ref{fig:different_binary} keeps the unary score fixed as the CLIP text-image similarity and shows the effects of varying the binary score function.}

\subsection{Dataset} \label{sec:sup_details_datasets}
\myparagraph{Text to image generation.} All text-to-image generation results with models FLUX.1 Dev, FLUX.1 Schnell, and SD3 (M) use all 553 prompts from the GenEval dataset~\cite{ghosh2023geneval}. 

\myparagraph{Depth to image generation.} All FLUX.1 Depth experiments use depth maps computed using Depth Anything Large ~\cite{depthanything} from 250 images from the validation split of the COCO 2017 dataset~\cite{lin2014microsoft}.  

\myparagraph{Encoder-based image customization.}
For all encoder-based image customization experiments using SynCD~\cite{kumari2025syncd}, we use 400 samples from the images in the standard DreamBooth dataset.

\begin{figure*}[t!]
    \centering 
    \includegraphics[width=\linewidth]{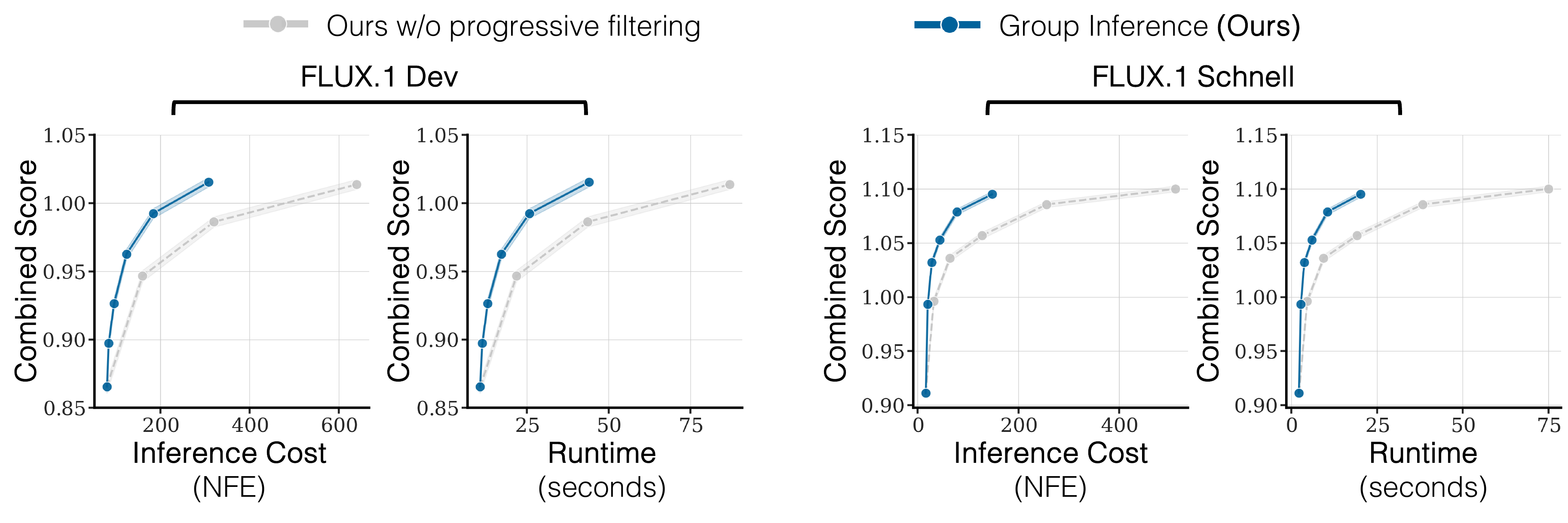}
    \vspace{-3mm}
    \caption{ 
    \textbf{Ablating the effect of progressive pruning.} Similar for Figure~\ref{fig:abl_progressive_filtering} from the main paper, we show the importance of progressive pruning. We report both, the number of function evaluations (NFEs) and the wallclock runtime (using one NVIDIA H100). The two plots on the left show the comparison using FLUX.1 Dev. The two plots on the right show FLUX.1 Schnell comparison.
    } 
    \label{fig:sup_ablation_pruning_nfe}
\end{figure*}

\begin{figure*}[t!]
    \centering 
    \includegraphics[width=0.7\linewidth]{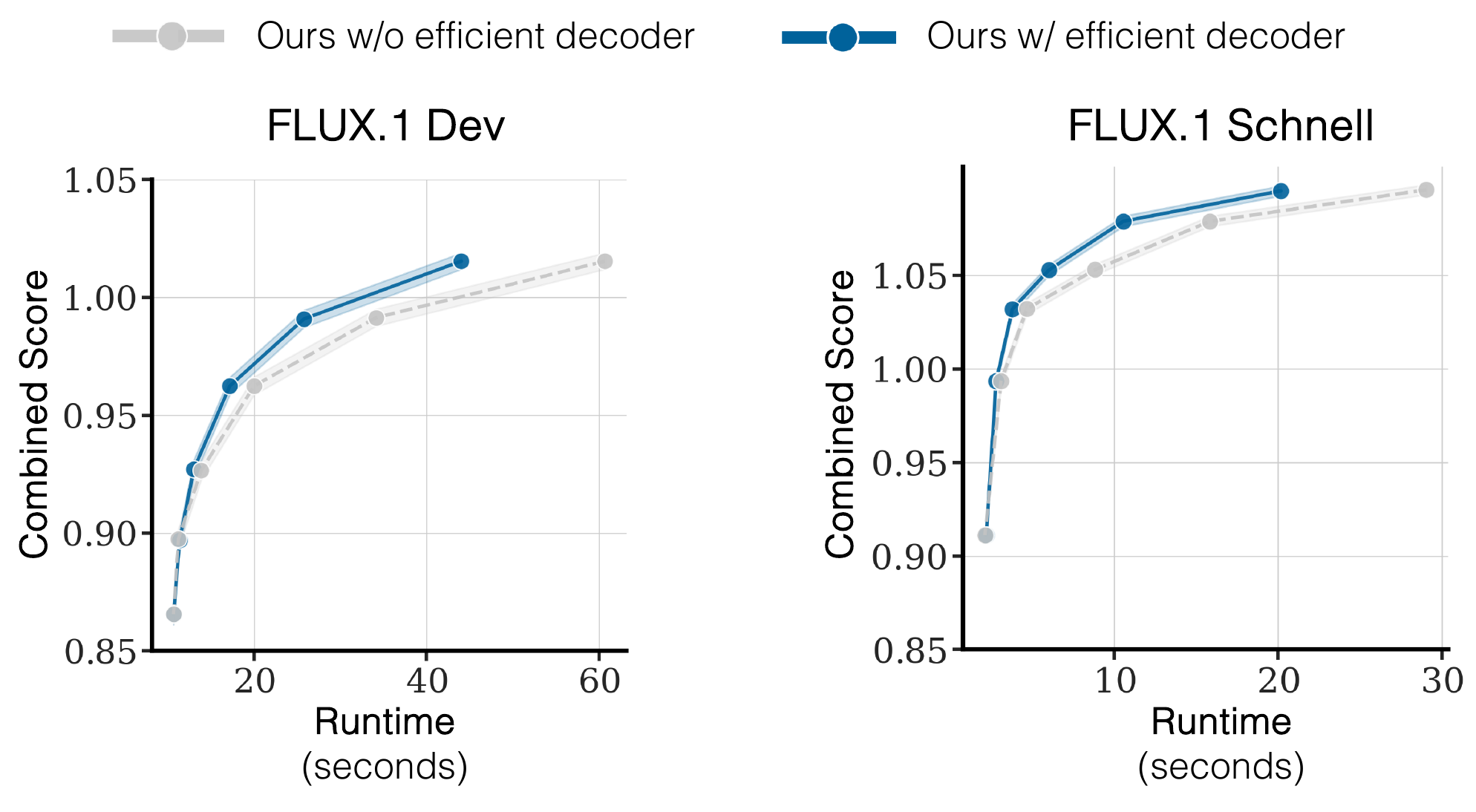}
    \caption{ 
    \textbf{Ablating efficient decoder.} We show the effects of using an efficient decoder for decoding the intermediate predictions. 
    } 
    \label{fig:sup_ablation_vae}
\end{figure*}

\begin{figure}[t!]
    \centering 
    \includegraphics[width=1.0\linewidth]{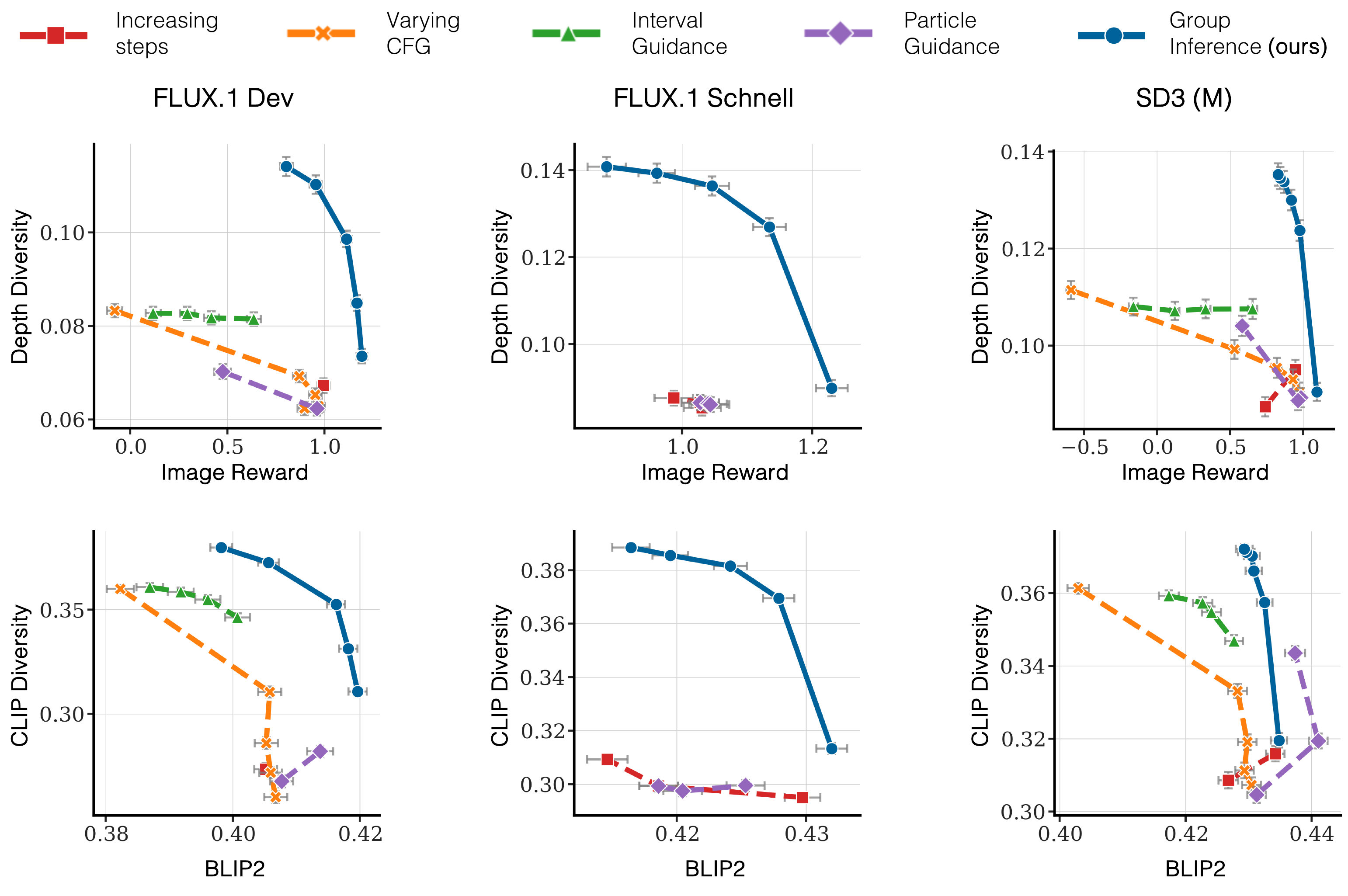}
    \caption{ \review{ \textbf{Quality and Diversity Pareto front with additional metrics.} We evaluate the quality and diversity of samples generated by different inference strategies for three text-to-image models (FLUX.1 Dev, FLUX.1 Schnell, and Stable Diffusion 3 Medium). The top row shows evaluation using Image Reward~\cite{xu2023imagereward} as the quality metric and Depth Diversity as the diversity metric. The bottom row uses BLIP2~\cite{li2023blip} and CLIP Diversity. Note that these metrics are unseen and not used by our method.
    }} 
    \label{fig:baselines_UB_other_metrics}
    \vspace{-3mm}
\end{figure}

\begin{figure*}[t!]
    \centering 
    \includegraphics[width=\linewidth]{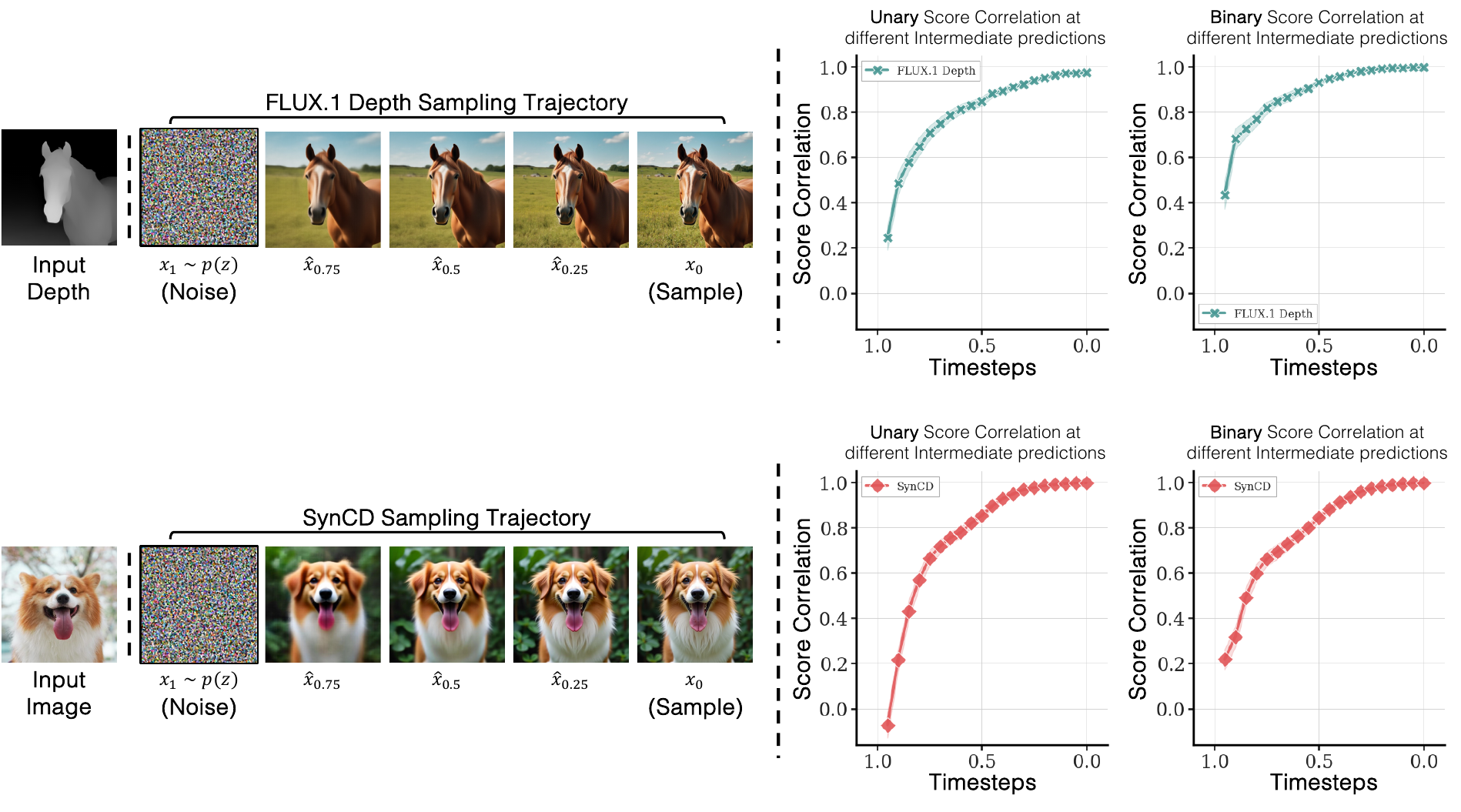}
    \caption{ 
    \textbf{Correlation between intermediate and final generation Scores .} We follow the same protocol as Figure 3 in the main paper. 
    On the left, we show the reverse diffusion process, visualizing the intermediate predictions $\hat{\x_t}$ of the final image at different steps for FLUX.1 Depth and SynCD models. We can observe that the intermediate predictions look similar to true final sample $\x_0$ for both the models.
    We further demonstrate this quantitatively by plotting the Spearman correlation of the Unary and Binary scores from $\hat{\x_t}$ versus final $\x_0$ scores, across different steps.
    } 
    \label{fig:sup_correlations_all}
\end{figure*}

\begin{figure*}[t!]
    \centering 
    \includegraphics[width=\linewidth]{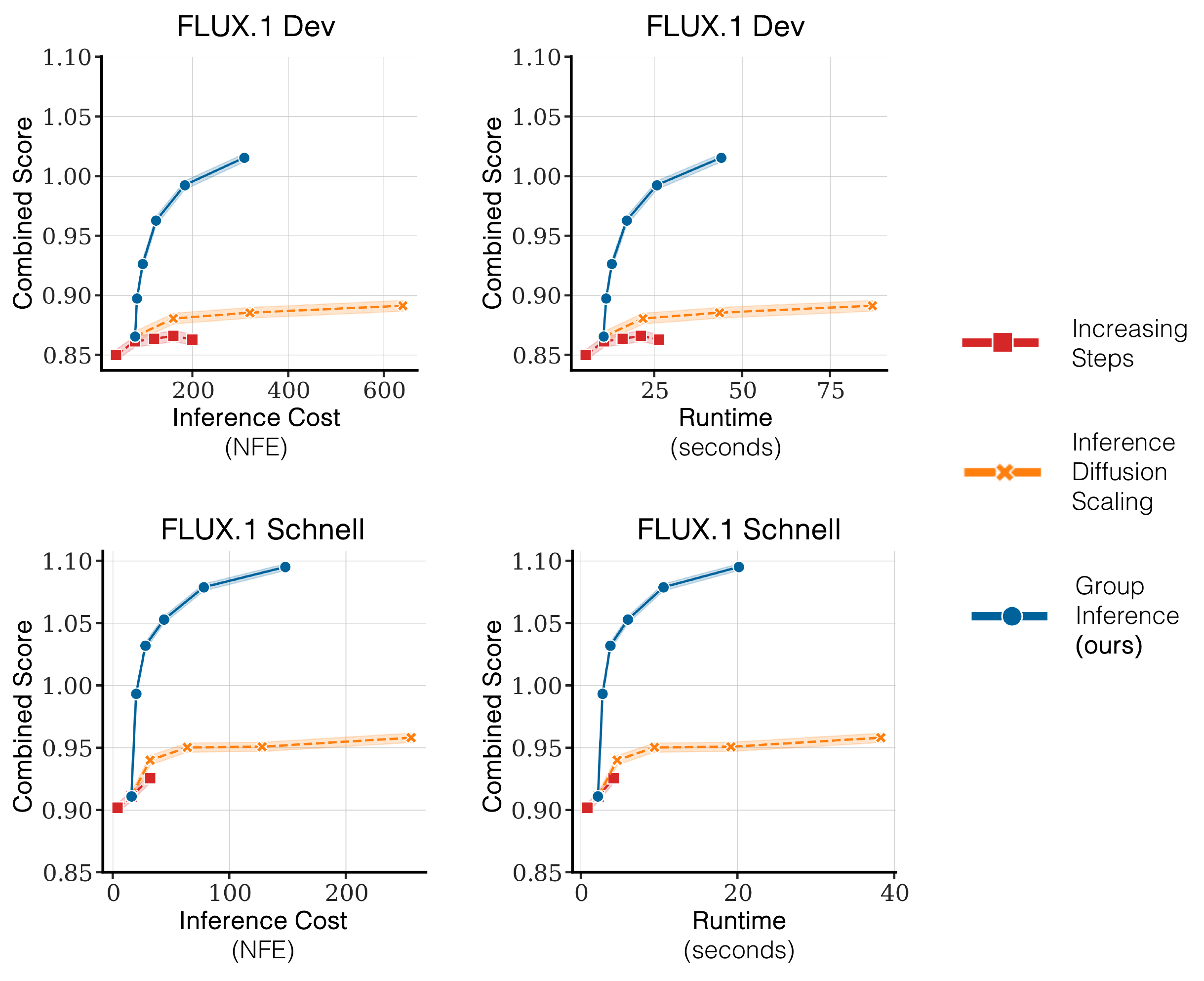}
    \vspace{-3mm}
    \caption{ 
    \textbf{Performance at different runtimes.} Similar for Figure 5 from the main paper, we show the different ways of allocating inference budget. We report both, the number of function evaluations (NFEs) and the wallclock runtime (using one NVIDIA H100). 
    } 
    \label{fig:sup_scaling_nfe}
\end{figure*}

\begin{figure}[t!]
    \centering 
    \includegraphics[width=\linewidth]{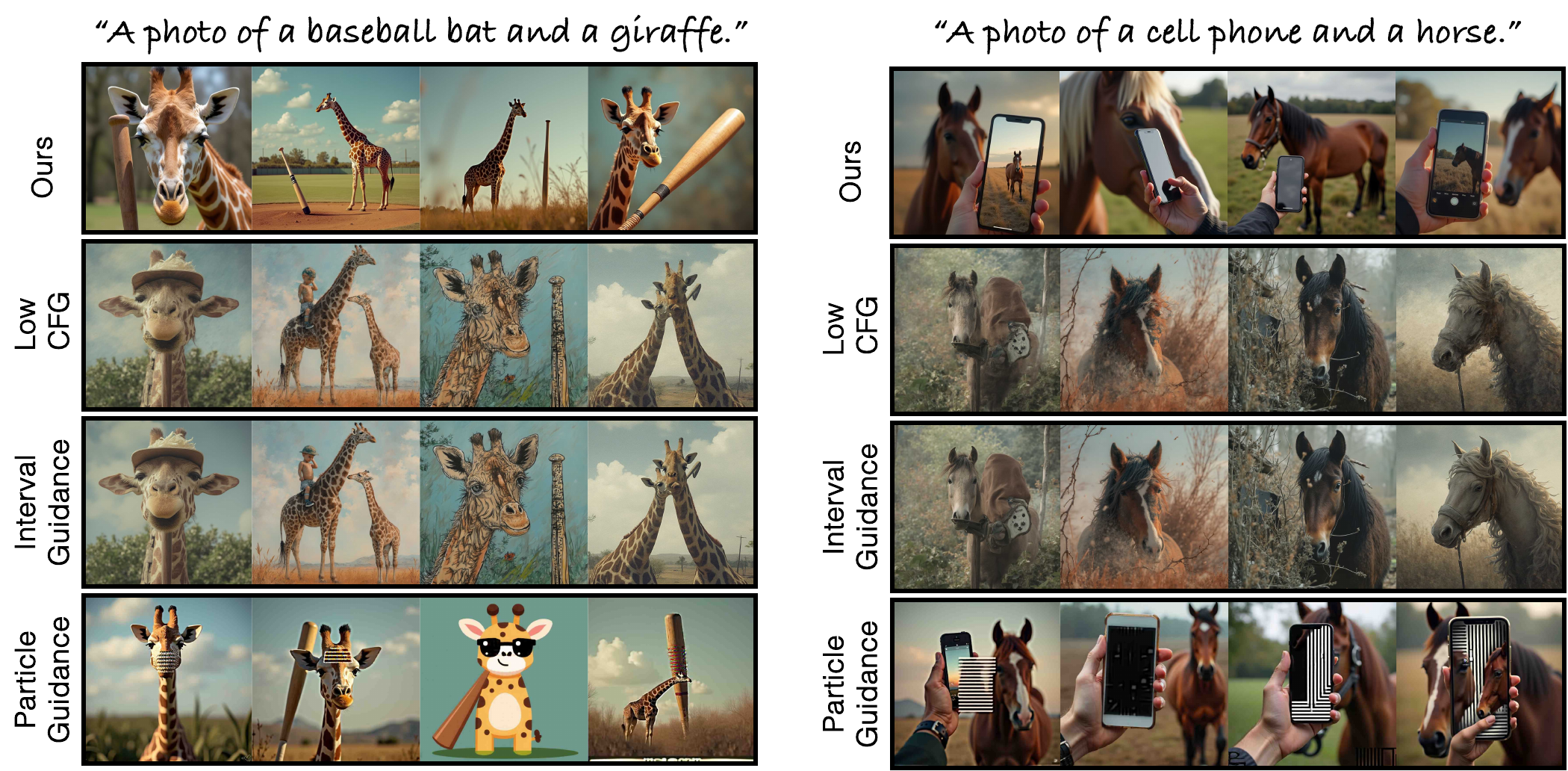}
    \caption{ \textbf{Qualitative results.} 
    We compare our proposed method (top row) against alternative inference strategies targeting an improved Quality-Diversity tradeoff with FLUX.1 Dev base model.
    } 
    \label{fig:supp_visual_comparison}
    \vspace{-4mm}
\end{figure}

\begin{figure*}[ht!]
    \centering 
    \includegraphics[width=\linewidth]{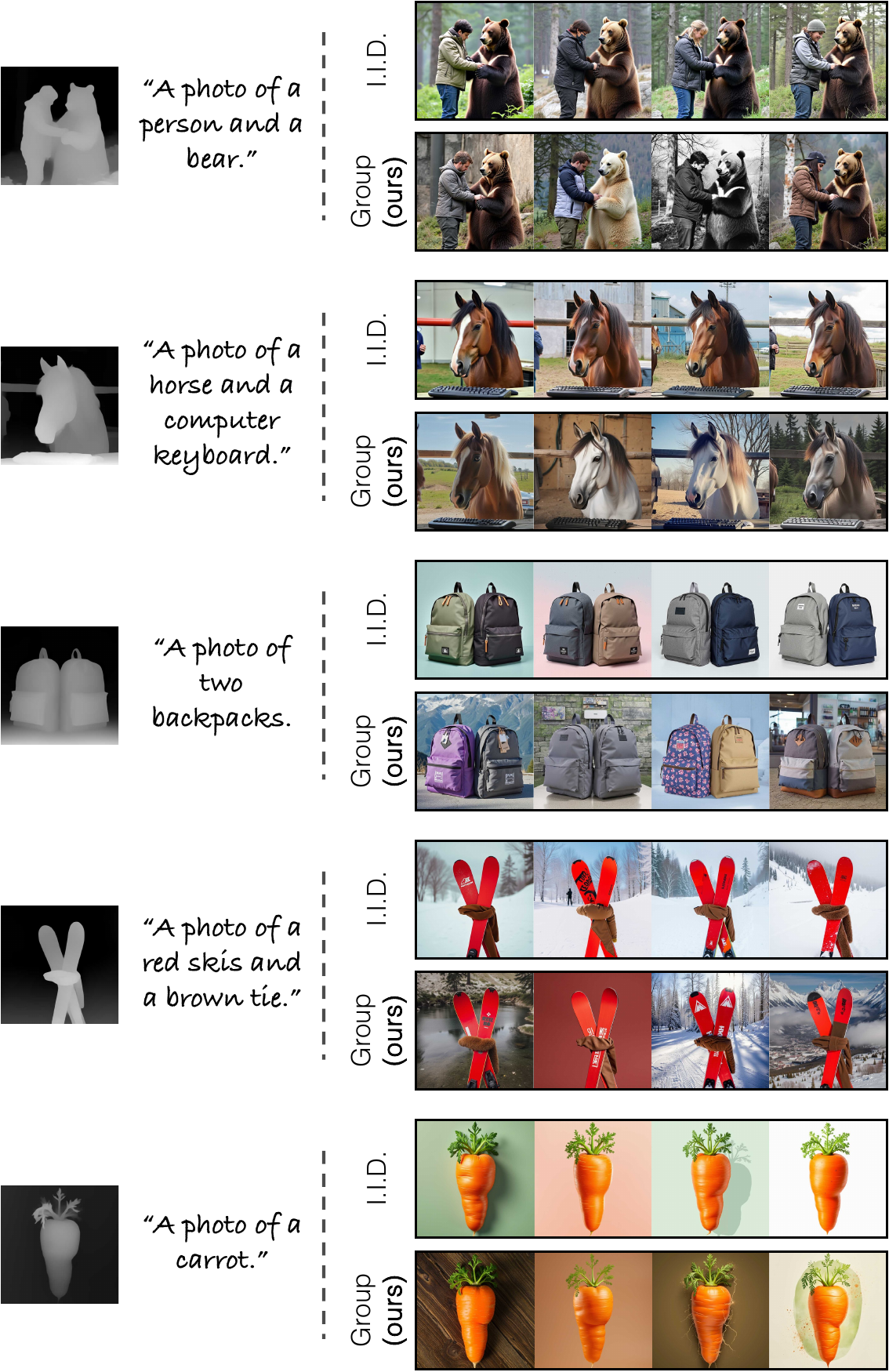}
    \caption{ \textbf{Gallery of results.} Qualitative results that show the advantage of our proposed method over I.I.D. sampling for depth-to-image generation using FLUX.1 Depth as the base model. The input depth maps and captions are shown on the left and the generated outputs are shown on the right. Our method consistently generates outputs that have more diverse backgrounds, styles, and textures. 
    } 
    \label{fig:supp_results_depth_a}
\end{figure*}

\begin{figure*}[ht!]
    \centering 
    \includegraphics[width=\linewidth]{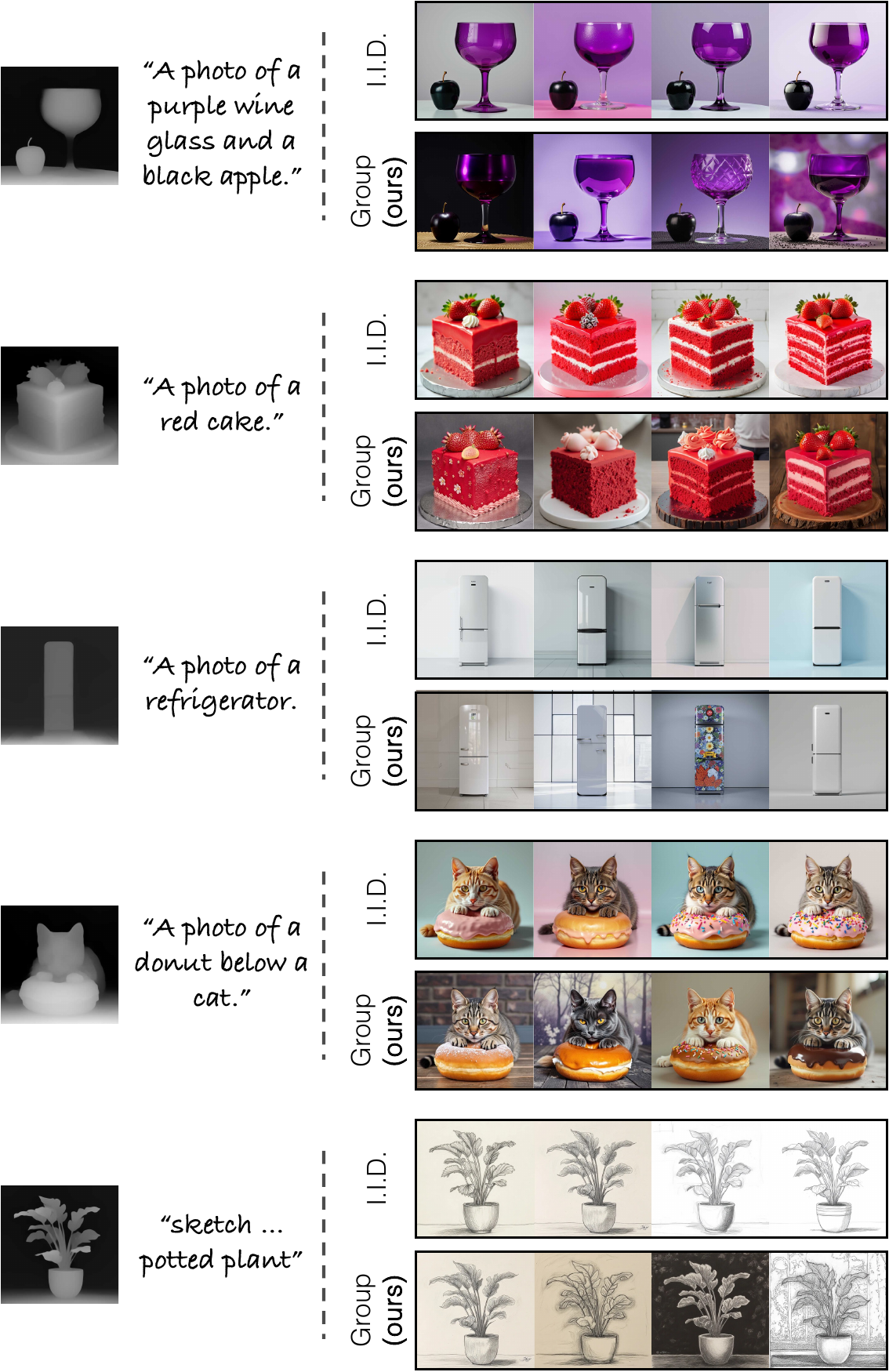}
    \caption{ \textbf{Gallery of results.} Qualitative results that show the advantage of our proposed method over I.I.D. sampling for depth-to-image generation using FLUX.1 Depth as the base model. The input depth maps and captions are shown on the left and the generated outputs are shown on the right. Our method consistently generates outputs that have more diverse backgrounds, styles, and textures. 
    } 
    \label{fig:supp_results_depth_b}
\end{figure*}

\begin{figure*}[ht!]
    \centering 
    \includegraphics[width=\linewidth]{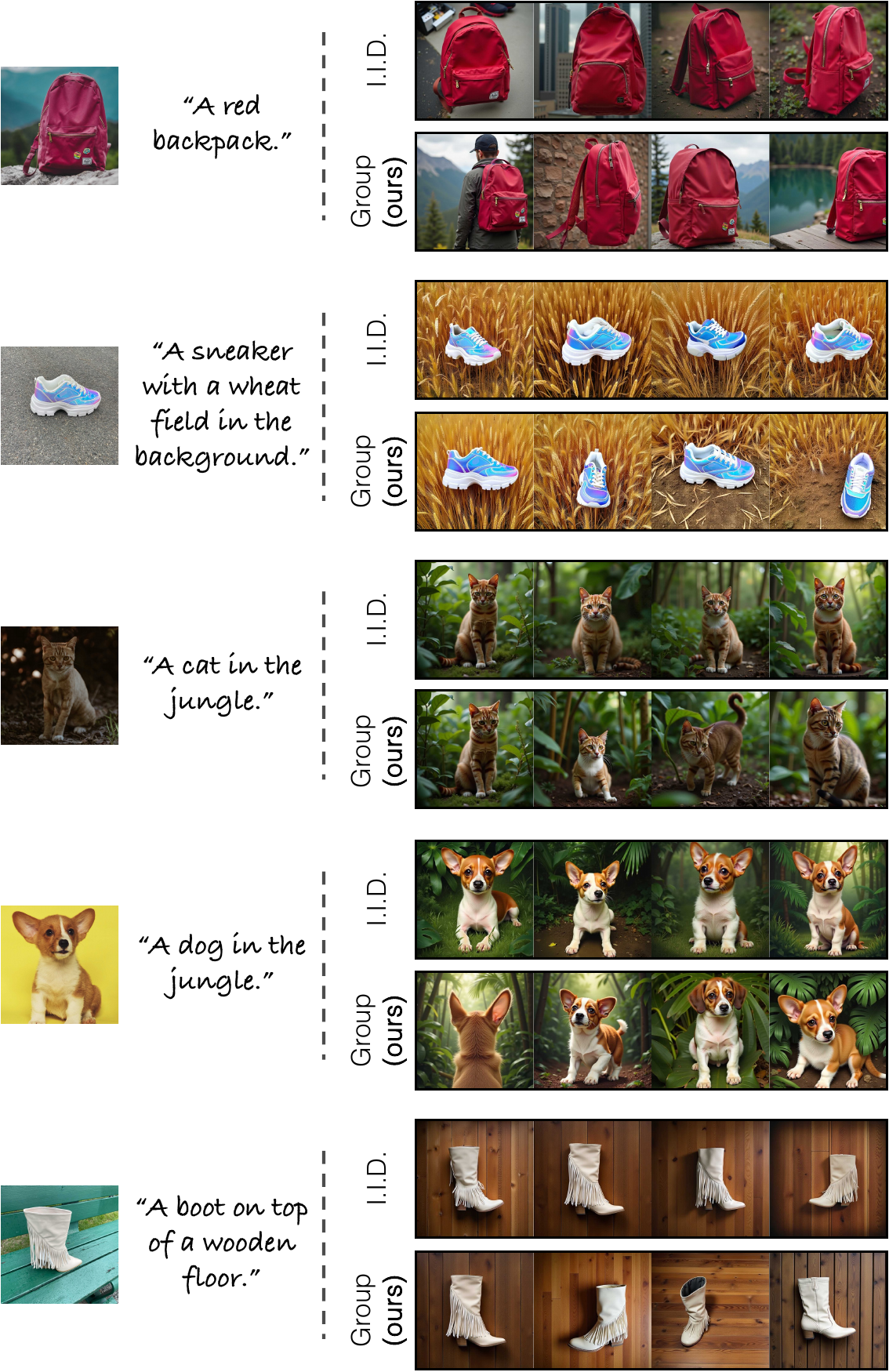}
    \caption{ \textbf{Gallery of results.} Qualitative results that show the advantage of our proposed method over I.I.D. sampling for feedforward customized generation using SynCD~\cite{kumari2025syncd}. The input image and captions are shown on the left and the generated outputs are shown on the right. Our method consistently generates outputs that have more diverse backgrounds, object poses, and styles. 
    } 
    \label{fig:supp_results_syncd_a}
\end{figure*}

\end{document}